\documentclass[10pt,twocolumn,letterpaper]{article}

\usepackage[pagenumbers]{iccv} 

\usepackage{float}
\usepackage[ruled,vlined]{algorithm2e}
\usepackage{algorithmic}
\usepackage{threeparttable}
\usepackage{multicol}
\usepackage{multirow}
\usepackage{booktabs} 
\usepackage{makecell} 
\usepackage{amsmath}
\usepackage{bm}
\usepackage{bbding}
\definecolor{iccvblue}{rgb}{0.21,0.49,0.74}
\usepackage[pagebackref,breaklinks,colorlinks,allcolors=iccvblue]{hyperref}

\def\paperID{5139} 
\def\confName{ICCV}
\def\confYear{2025}

\title{Meta Curvature-Aware Minimization for Domain Generalization}

\author{
Ziyang Chen$^{1}$~~~Yiwen Ye$^{1}$~~~Feilong Tang$^{2}$~~~Yongsheng Pan$^{1\dag}$~~~Yong Xia$^{1, 3, 4\dag}$\\
$^{1}$ School of Computer Science and Engineering, Northwestern Polytechnical University, China\\
$^{2}$ Faculty of Engineering, Monash University, Australia\\
$^{3}$ Research \& Development Institute of Northwestern Polytechnical University in Shenzhen, China\\
$^{4}$ Ningbo Institute of Northwestern Polytechnical University, China\\
{\{zychen, ywye\}@mail.nwpu.edu.cn,
feilong.tang@monash.edu,
\{yspan, yxia\}@nwpu.edu.cn
}
}

\begin{document}
\maketitle



\begin{abstract}
Domain generalization (DG) aims to enhance the ability of models trained on source domains to generalize effectively to unseen domains. Recently, Sharpness-Aware Minimization (SAM) has shown promise in this area by reducing the sharpness of the loss landscape to obtain more generalized models. However, SAM and its variants sometimes fail to guide the model toward a flat minimum, and their training processes exhibit limitations, hindering further improvements in model generalization.
In this paper, we first propose an improved model training process aimed at encouraging the model to converge to a flat minima.
To achieve this, we design a curvature metric that has a minimal effect when the model is far from convergence but becomes increasingly influential in indicating the curvature of the minima as the model approaches a local minimum.
Then we derive a novel algorithm from this metric, called \textbf{Me}ta \textbf{C}urvature-\textbf{A}ware \textbf{M}inimization (MeCAM), to minimize the curvature around the local minima.
Specifically, the optimization objective of MeCAM simultaneously minimizes the regular training loss, the surrogate gap of SAM, and the surrogate gap of meta-learning.
We provide theoretical analysis on MeCAM's generalization error and convergence rate, and demonstrate its superiority over existing DG methods through extensive experiments on five benchmark DG datasets, including PACS, VLCS, OfficeHome, TerraIncognita, and DomainNet. Code will be available on GitHub.
\end{abstract}

\renewcommand{\thefootnote}{}
\footnote{$^\dag$Corresponding author. 
}

\begin{figure}[t]
  \centering
  \includegraphics[width=\columnwidth]{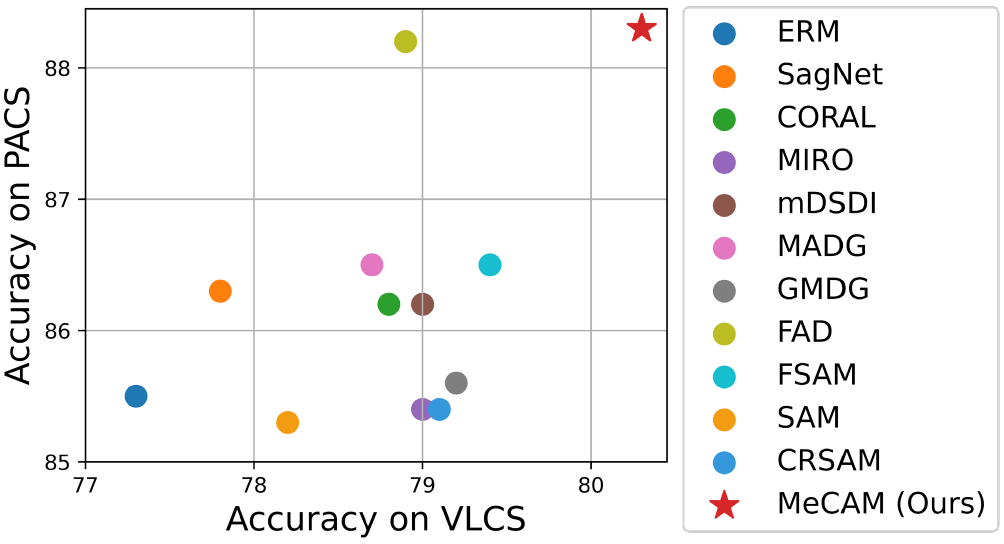}
   \caption{Accuracy of our MeCAM and existing DG methods on the PACS and VLCS datasets. MeCAM achieves superior generalization performance on both datasets.} 
\label{fig:accuracy}
\end{figure}

\section{Introduction}
\label{sec:intro}
In recent years, deep neural networks (DNNs) have demonstrated significant success across various computer vision tasks~\cite{DNNsurvey_1,DNNsurvey_2,DNNsurvey_3}. However, models trained on source data often experience performance degradation when applied to unseen target domains due to distribution shifts~\cite{domain_shift_cvpr}. This limitation hinders the practical deployment of these models in real-world scenarios.

Domain generalization (DG) has emerged as a promising approach to address this challenge by leveraging source data to train models that generalize well to new, unseen domains. Several DG techniques have been developed to improve model generalization, including domain alignment~\cite{alignment_1,alignment_2}, data augmentation~\cite{augmentation_2,augmentation_3}, ensemble learning~\cite{ensemble_1,ensemble_2}, and disentangled representation learning~\cite{disentangled_1,disentangled_2}, and so on.

Despite these advancements, simply minimizing standard loss functions, such as cross-entropy loss, often fails to achieve satisfactory generalization~\cite{SAM}, as it may result in a sharp loss landscape. Recent research~\cite{flatminima_1,flatminima_2,flatminima_3} has explored the relationship between the geometry of the loss landscape and model generalization. Sharpness-aware minimization (SAM) and its variants~\cite{SAM,SAGM,GSAM,FSAM,CRSAM} have shown that reducing the sharpness of the loss landscape leads to improved generalization. These sharpness-based methods generally optimize the model by perturbing its parameters within a small radius, aiming to find flatter regions in the loss landscape.

While SAM has been successful in achieving flatter loss landscapes, it exhibits limitations in representing sharpness. Specifically, SAM measures sharpness based on the loss value, whereas sharpness itself is independent of the loss value. Moreover, SAM aims to reduce sharpness consistently, even when the model is not yet close to a local minimum. We argue that the model training process should follow two key principles: (1) focus on minimizing the loss when the model is far from convergence, and (2) shift the focus to reducing sharpness as the model approaches a local minimum. This approach is more reasonable, as unconverged models do not require sharpness consideration, and sharpness should not be dependent on the loss value.

To address these issues, we first introduce a curvature metric that adheres to these principles. This metric is independent of the loss value and effectively measures the curvature around local minima. As the model converges, the metric increases, reflecting the curvature of the loss landscape. Our goal is to minimize both this curvature metric and the vanilla training loss to train a model with better generalization ability.
Building on this, we propose a novel sharpness-based algorithm, Meta Curvature-Aware Minimization (MeCAM), which jointly minimizes the surrogate gap of SAM and meta-learning to find a flatter minima for better generalization. The optimization objective of MeCAM incorporates the vanilla training loss, the surrogate gap of SAM, and the surrogate gap of meta-learning. Minimizing the training loss leads the model toward a local minimum, while minimizing the other two terms reduces the curvature, thereby improving generalization. MeCAM encourages the model to converge to a flatter region, enhancing its generalization ability (see Figure~\ref{fig:accuracy}).

The contributions of this work are three-fold.
(1) We analyze the limitations of SAM and introduce principles for a more effective training process. We propose a curvature metric that better represents sharpness than SAM.
(2) We introduce the MeCAM algorithm, which jointly minimizes the surrogate gap of SAM and meta-learning to find a flatter minima.
(3) Extensive experiment results on five DG datasets demonstrate that MeCAM outperforms existing DG methods, achieving superior generalization performance.

\section{Related Work}
\label{sec:work}

\subsection{DG Methods}
DG methods~\cite{dg_survey1,dg_survey2} aim to train models on source data in a way that enables them to generalize well to other out-of-distribution data. Existing DG approaches can be broadly categorized into four groups based on their methodology and motivation: (1) domain alignment~\cite{alignment_1,alignment_2,alignment_3,alignment_4,alignment_5}, which measures the distance between distinct distributions and learns domain-invariant representations to enhance the robustness of the model; (2) data augmentation~\cite{MixStyle,augmentation_2,augmentation_3,augmentation_4}, which strongly enhances the diversity of training data to prevent the model from over-fitting to the training data; (3) ensemble learning~\cite{ensemble_1,ensemble_2,ensemble_3,ensemble_4}, which trains multiple models and uses their ensemble to improve predictions and reduce bias; and (4) disentangled representation learning~\cite{disentangled_1,disentangled_2,disentangled_3}, which separates features into domain-variant and domain-invariant components, using the domain-invariant features for more robust predictions. 
Although these DG methods have achieved significant success in improving model generalization, they may still underperform on out-of-distribution data due to a lack of constraints on the sharpness of the loss landscape. In this paper, we introduce a novel sharpness-based approach to guide the model toward a flatter minima, further enhancing generalization.

\subsection{SAM in DG}
The connection between the sharpness of the loss landscape and generalization ability has been explored in earlier studies~\cite{flatminima_1,flatminima_2,flatminima_3,flatminima_4}. 
For instance, Dinh \emph{et al.}~\cite{flatminima_4} suggested that the sharpness of a local minima is related to the Hessian spectrum, with the eigenvalues of the Hessian matrix reflecting the curvature of the loss landscape. Recent advancements in SAM and its variants~\cite{SAM,GSAM,GAM,CRSAM,SAGM,ESAM,ES-SAM,ASAM} have shown that flatter minima are associated with better generalization. SAM~\cite{SAM,understand_sam} uses worst-case perturbations of model weights to force the model to converge to a flatter minima, thereby improving generalization. Zhuang \emph{et al.}~\cite{GSAM} and Wang \emph{et al.}~\cite{SAGM} focus on optimizing the gap between the perturbation loss in SAM and the vanilla training loss. 
Wu \emph{et al.}~\cite{CRSAM} present the normalized Hessian trace as a regularizer for SAM to counter excessive non-linearity of loss landscape and compute the trace via finite differences.
In contrast, our approach derives a new optimization objective from a curvature metric, which offers a more precise measure of the sharpness around the local minima. 
We then derive the optimization objective of jointly minimizing the surrogate gap of SAM and meta-learning from minimizing this metric, facilitating the search for a flatter minima with better generalization performance.

\subsection{Meta-learning in DG}
Meta-learning~\cite{meta_survey2,meta_survey3}, also known as learning-to-learn, focuses on leveraging knowledge from related tasks to improve learning on new tasks. Recent studies have introduced meta-learning techniques into the domain generalization setting, enabling models to generalize across diverse domains by identifying patterns that are transferable to unseen domains. Finn \emph{et al.}~\cite{MAML} proposed a meta-learning approach that divides the training data into meta-train and meta-test sets, training the model to boost its performance on the meta-test set. Li \emph{et al.}~\cite{MLDG} introduced a model-agnostic training procedure that simulates distribution shifts by synthesizing virtual test domains. Balaji \emph{et al.}~\cite{Metareg} applied meta-learning by encoding domain generalization as a regularization function, learning a generalized regularizer to guide the training process. In this paper, we propose MeCAM, which enhances the conventional meta-learning with SAM to guide the model toward a flatter minima, thereby further improving its generalization performance.

\section{Methodology}
\label{sec:method}
\subsection{Preliminaries}
\subsubsection{Problem Definition}
Let $\theta \in \mathbb{R}^k$ represent the model parameters, where $k$ is the parameter dimension. We denote the training datasets as $\mathcal{D} = \{ \mathcal{D}_i \}_i^S$, consisting of $S$ datasets, each containing $n_i$ training samples, denoted by $\{ (x_i^j, y_i^j)\}^{n_i}_{j=1} \sim \mathcal{D}_i$, where $x_i^j$ and $y_i^j$ are the input data and corresponding target labels, respectively. The training loss function over the dataset $\mathcal{D}$ is defined as $f(\theta;\mathcal{D})$, based on cross-entropy loss. For simplicity, we denote $f(\theta;\mathcal{D})$ as $f(\theta)$ unless otherwise specified. The first derivative and the Hessian matrix of $f(\theta)$ at point $\theta$ are denoted by $\nabla f(\theta)$ and $H(\theta)$, respectively. We use $\| \cdot \| $ to represent the L2 norm throughout the paper.

\subsubsection{Objective of SAM}
Conventional optimization of DNNs typically minimizes the training loss $f(\theta)$ using gradient descent, but this often leads to sharp local minima, which can hinder generalization performance~\cite{GSAM}. To address this issue, SAM~\cite{SAM} aims to find a flatter region in the loss landscape where the model parameters are robust to small perturbations. The objective of SAM can be formulated as:
\begin{equation}
\begin{aligned}
\min_\theta f(\theta+\delta), \,\,  \delta = \rho \frac{\nabla f(\theta)}{\| \nabla f(\theta)\| + \epsilon}
\label{eq:sam}
\end{aligned}
\end{equation}
where $\delta$ denotes the perturbation, $\rho$ controls the perturbation radius, determining the amplitude of the perturbation, and $\epsilon$ is a small constant to avoid division by zero.

\subsection{Curvature Metric}
\begin{figure*}[!t]
  \centering
  \includegraphics[width=\textwidth]{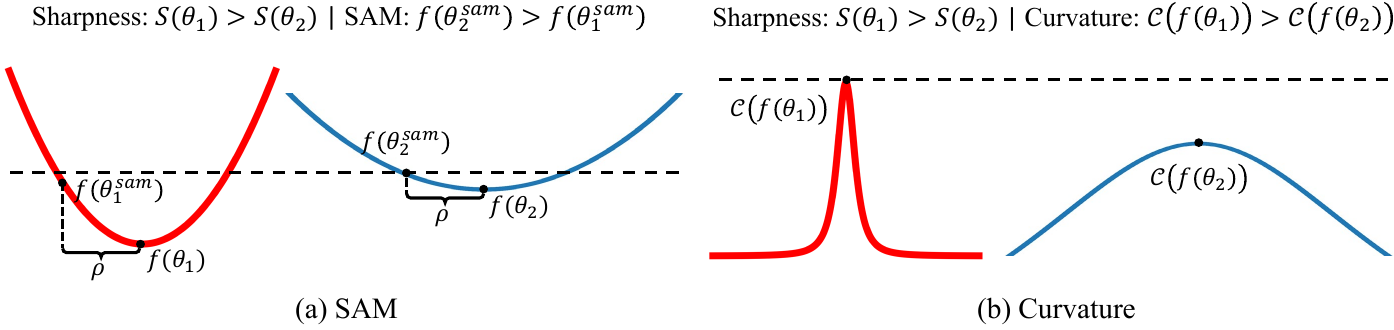}
   \caption{Comparison between SAM and the proposed curvature metric. (a) Illustration of a sharp local minimum at $\theta_1$ (red) and a flat local minimum at $\theta_2$ (blue), where $S(\theta)$ denotes the sharpness of model with parameter $\theta$. While SAM prefers to optimize around $\theta_2^{sam}$, $\theta_2$ is inherently flatter than $\theta_1$. This highlights SAM's limitation in accurately characterizing sharpness, as it is influenced by the loss value. (b) Illustration of the proposed curvature metric $\mathcal{C}$. Unlike SAM, $\mathcal{C}$ more effectively quantifies the deviation from a flat surface, providing a better measure of sharpness in the loss landscape. A smaller value of $\mathcal{C}$ indicates a flatter minimum. Best viewed in color.} 
\label{fig:curvature}
\end{figure*}

Since SAM measures sharpness based solely on the loss value, it may fail to accurately capture the true sharpness in some situations~\cite{GAM,GSAM,SAGM}. 
As shown in Figure~\ref{fig:curvature} (a), a smaller $f(\theta^{sam})$ does not always guarantee a flatter minima. Additionally, SAM consistently minimizes sharpness during training, without considering that the model should focus on minimizing the loss during early stages of training and only reduce sharpness near local minima.
We propose a more rational training process that follows two principles: (1) minimize the loss when the model is far from convergence; and (2) reduce the sharpness when the model approaches a local minimum.
Inspired by~\cite{federer1959curvature}, we define a curvature metric that is independent of the loss value and adheres to these principles, providing a more accurate measure of sharpness and improving the model's training process. The curvature metric is formulated as:
\begin{equation}
\begin{aligned}
\mathcal{C}(f(\theta)) = \frac{\|H(\theta)\|}{\| \nabla f(\theta) \|^2+1}.
\label{eq:curvature}
\end{aligned}
\end{equation}
This metric plays a minor role when the model has not converged but becomes more significant as the model approaches a local minimum, as shown in Figure~\ref{fig:curvature} (b).
We also explore the relationship between this metric and the maximal eigenvalue of the Hessian, $\lambda_{max}(H(\theta))$, which is a proper measure of the sharpness of the local minima~\cite{kaur2023maximum,flatminima_3} and is related to the generalization ability of the model~\cite{chen2022vision}.
Although $\lambda_{max}(H(\theta))$ is challenging to approximate and optimize directly~\cite{GAM,yao2020pyhessian}, we derive the following lemma showing that our curvature metric can be served as a suitable surrogate for it.

\noindent{\textbf{Lemma 3.1.}} \emph{For a local minima, assuming that $H(\theta)$ is positive semi-definite~\cite{GSAM}, we have}
\begin{equation}
\begin{aligned}
\lambda_{max}(H(\theta)) = \mathcal{C}(f(\theta)).
\label{eq:lemma32}
\end{aligned}
\end{equation}
This indicates that $\lambda_{max}(H(\theta))$ can be directly controlled by our curvature metric.



\subsection{Meta Curvature-Aware Minimization}
As $\theta$ approaches a local minimum, the gradient $\nabla f(\theta)$ in Eq. (\ref{eq:curvature}) inevitably decreases. Consequently, we minimize the Hessian to reduce the numerator approximatively.
Based on this analysis, we propose the MeCAM framework, which optimizes the following overall objective:
\begin{equation}
\begin{aligned}
\min_\theta f(\theta), H(\theta).
\label{eq:overall_obj}
\end{aligned}
\end{equation}
Here, minimizing $f(\theta)$ helps $\theta$ approach the local minima, and minimizing $H(\theta)$ focuses on reducing the curvature around the local minima.

Calculating $H(\theta)$ for a large matrix, such as a DNN with millions of parameters, is computationally expensive. To reduce complexity, we approximate $H(\theta)$ using the central difference method:

\noindent{\textbf{Theorem 3.2.}} \emph{Suppose $f(\theta)$ is twice-differentiable at $\theta$. Using the central difference method, we have}
\begin{equation}
\begin{aligned}
H(\theta) \approx \frac{f(\theta+h)+f(\theta-h)-2f(\theta)}{h^2},
\label{eq:theorem33}
\end{aligned}
\end{equation}
where $h$ is a small step-size, and the remainder term is omitted.
We estimate $H(\theta)$ by calculating finite differences at symmetric neighboring points around the point of interest.
By incorporating information from both sides, this approach can provide an accurate approximation of $H(\theta)$.
A key consideration in Eq. (\ref{eq:theorem33}) is the choice of $h$.
Common strategies for choosing an appropriate value of $h$ include balancing round-off errors, adjusting based on the scale of the problem, using typical small constants, and so on~\cite{nocedal1999numerical,dennis1996numerical}. 
To avoid additional computation, in this study, we empirically set 
\begin{equation}
\begin{aligned}
h=\delta=&\rho \frac{ \nabla f(\theta)}{\| \nabla f(\theta) \| + \epsilon}
\end{aligned}
\end{equation}
and utilize a hyperparameter $\alpha$ to replace $\frac{1}{h^2}$ to better control the contribution of this term during training.
The objective in Eq. (\ref{eq:overall_obj}) can then be rewritten as:
\begin{equation}
\begin{aligned}
\min_\theta f(\theta) + \alpha(f(\theta+\delta)-f(\theta)) + \alpha(f(\theta-\delta)-f(\theta)).
\label{eq:overall_obj2}
\end{aligned}
\end{equation}
When the model has not converged, assuming $f$ is monotonic and smooth within $[\theta+\delta, \theta-\delta]$ and $\delta$ is small, the sum of the second and third items in Eq.~\ref{eq:overall_obj2} tends to $0$.
Conversely, as the model approaches a local minimum, this sum remains nonnegative and is positively correlated with the model's sharpness.

\noindent{\textbf{Generalization analysis.}}
Here we derive a generalization bound based on PAC-Bayesian theorem~\cite{PAC-Bayesian} for our MeCAM in Proposition~\ref{proposition_33}.

\noindent{\textbf{Proposition 3.3.} 
\label{proposition_33}
\emph{Suppose the loss function $f$ is differentiable and bounded by $M$, and the training set consists of $n$ i.i.d. samples drawn from the true distribution. Let $\hat{f}(\theta)$ represent the loss on the training set. Let $\theta \in \mathbb{R}^k $ be learned from the training set with a number of $k$. Then, with probability at least $1-\zeta$, we have}
\begin{equation}
\begin{aligned}
&\mathbb{E}_{\delta_i \sim \mathcal{N}(0,\sigma^2)}[f(\theta+\delta )] \le
\hat{f}(\theta) + \\
&\alpha(\hat{f}(\theta+\delta)+\hat{f}(\theta-\delta)-2\hat{f}(\theta)) + \frac{M}{\sqrt{n}} + \\
&\sqrt{\frac{\frac{1}{4}k\log (1+\frac{\| \theta \|^2 (1+\sqrt{\frac{\ln n}{k}})^2}{\rho^2})+\frac{1}{4}+\log \frac{n}{\zeta}+2\log(6n+3k)}{n-1}}.
\label{eq:proposition_43}
\end{aligned}
\end{equation}
It implies that $f(\theta)$ is bounded by $\hat{f}(\theta) + \alpha(\hat{f}(\theta+\delta)+\hat{f}(\theta-\delta)-2\hat{f}(\theta))$ when ignoring high-order terms, and minimizing the second term in the right side of Eq. (\ref{eq:proposition_43}) is expected to tighten the upper bound, rendering $\alpha(\hat{f}(\theta+\delta)+\hat{f}(\theta-\delta)-2\hat{f}(\theta))$ a regularization term controlled by the hyperparameter $\alpha$.

\

\noindent{\textbf{Convergence analysis.}} We analyze the convergence rate of MeCAM under the assumption that $f$ is $L$-Lipschitz smooth.

\noindent{\textbf{Theorem 3.4.} 
\emph{Suppose $f(\theta)$ is $L$-Lipschitz smooth. For any time-step $t$ and $\theta \in \Theta$, suppose we can observe $g_t$, $g_t^{sam}$, and $g_t^{meta}$ as the gradients of $\nabla f(\theta_t)$, $\nabla f(\theta_t+\delta)$, and $\nabla f(\theta_t-\delta)$, with $\| g_t \|, \| g_t^{sam} \|, \| g_t^{meta} \| \le G$. Then, with learning rate $\eta_t = \frac{\eta_0}{\sqrt{t}}$ and perturbation radius $\rho_t = \frac{\rho_0}{\sqrt{t}}$, we have}
\begin{equation}
\begin{aligned}
\frac{1}{T} \sum_{t=1}^T \| \nabla f(\theta_t) \|^2 \le 
\frac{C_1 + C_2 \log T}{\sqrt{T}}, \\
\frac{1}{T} \sum_{t=1}^T \| \nabla f(\theta_t+\delta_t) \|^2 \le 
\frac{C_3 + C_4 \log T}{\sqrt{T}}, \\
\frac{1}{T} \sum_{t=1}^T \| \nabla f(\theta_t-\delta_t) \|^2 \le 
\frac{C_5 + C_6 \log T}{\sqrt{T}},
\label{eq:theorem_34}
\end{aligned}
\end{equation}
where $C_1$, $C_2$, $C_3$, $C_4$, $C_5$, and $C_6$ are some constants.
This implies that $f(\theta)$, $f(\theta+\delta)$, and $f(\theta-\delta)$ in MeCAM all converge at a rate of $O(\frac{\log T}{\sqrt{T}})$ for non-convex stochastic optimization, aligning our MeCAM with the convergence rate of first-order gradient methods, such as Adam.

As the third term in Eq. (\ref{eq:overall_obj2}) is a meta-like objective function, we further approximate our MeCAM as a meta-driven approach.
We perform perturbations on $\mathcal{D}$ to generate $\Tilde{\mathcal{D}}$, a virtual meta-test set. 
In this study, we apply Mixstyle~\cite{MixStyle} for perturbations. We then derive an approximation for the third term in Eq. (\ref{eq:overall_obj2}).

\SetKwInOut{Require}{Require}
\SetKwInOut{Initialize}{Initialize}
\begin{algorithm}[t]
\caption{MeCAM Algorithm.}
\small
\Initialize{Initialize $t$ and $\theta_0$ with $0$ and initial parameters, respectively.}
\KwIn{Training set $\mathcal{D}$, batch size $b$, learning rate $\eta_t$, perturbation radius $\rho_t$, hyperparameters $\alpha$ and $\beta$, small constant $\epsilon$, and total number of iterations $T$}
\begin{algorithmic}[1]
\FOR{$t$ \textbf{to} $T$}
\STATE Sample mini-batch data $\mathcal{B}$ from $\mathcal{D}$;
\STATE Forward $f(\theta_t)$ on $\mathcal{B}$ and then back-propagate the gradients;
\STATE Compute the perturbation $\delta_t = \rho_t \frac{\nabla f(\theta_t)}{\| \nabla f(\theta_t) \| + \epsilon}$;
\STATE Forward $f(\theta_t+\delta_t)$ on $\mathcal{B}$ and then back-propagate the gradients;
\STATE Forwad $m(\theta_t-\delta_t)$ on $\Tilde{\mathcal{B}}$ obtained by utilizing Mixstyle and then back-propagate the gradients;
\STATE Update: $\theta_{t+1} \leftarrow \theta_t - \eta_t((1-\alpha-\beta)\nabla f(\theta_t)+$ 
$\alpha \nabla f(\theta_t+\delta_t)+\beta \nabla m(\theta_t-\delta_t))$
\ENDFOR
\end{algorithmic}
\KwOut{$\theta_t$}
\label{algorithm1}
\end{algorithm}

\noindent{\textbf{Proposition 3.5.} 
\label{proposition_35}
\emph{When the perturbation $\delta$ is small, we have}
\begin{equation}
\begin{aligned}
\alpha (f(\theta-\delta) - f(\theta)) \approx \beta (m(\theta-\delta) - f(\theta)), 
\label{eq:meta_approx}
\end{aligned}
\end{equation}
where $\beta$ is another hyperparameter balancing the terms, with the condition of $\beta \le \alpha$. The proof can be found in Supplementary Material A.
Since we denote $f(\theta;\mathcal{D})$ as $f(\theta)$, we utilize $m(\theta-\delta)$ to represents $f(\theta-\delta;\Tilde{\mathcal{D}})$ for clarity.
$m(\theta-\delta)$ is designed to emulate the deployment environment to assess the model when transferred from the training set $\mathcal{D}$ to the virtual meta-test set $\Tilde{\mathcal{D}}$, indicating the robustness of the current model across different data distributions.
Finally, by combining both Eq. (\ref{eq:overall_obj2}) and Eq. (\ref{eq:meta_approx}), the overall objective of MeCAM can be defined as:
\begin{equation}
\begin{aligned}
&\min_\theta f(\theta) + \alpha(\underbrace{f(\theta+\delta)-f(\theta)})+\beta(\underbrace{m(\theta-\delta)-f(\theta)}),\\
&\quad surrogate \,\, gap \,\, of \,\, \underline{SAM} \,\, and \,\, \underline{meta\text{-}learning},
\label{eq:final_overall_obj}
\end{aligned}
\end{equation}
where the second and third terms in the right side can be regarded as the surrogate gap of SAM and the surrogate gap of meta-learning, respectively.
A smaller gap between the loss of SAM (\emph{i.e.}, $f(\theta+\delta)$) or meta-learning (\emph{i.e.}, $m(\theta-\delta)$) and the vanilla training loss (\emph{i.e.}, $f(\theta)$) indicates a flatter loss landscape, which contributes to the robustness of the model.
Previous work~\cite{GSAM,SAGM} has also demonstrated the superiority of the surrogate gap over the vanilla training loss. 
Compared to directly calculating the Hessian, the process in our MeCAM is much easier and less computational, allowing our method to be used like other conventional optimizers, such as Adam~\cite{Adam}.
The optimization procedure of MeCAM is outlined in Algorithm~\ref{algorithm1}.

\section{Experiments}
\label{sec:exper}

\begin{table}[!t]
\caption{The dataset details of PACS, VLCS, OfficeHome, TerraIncognita, and DomainNet.}
    \centering
    \resizebox{\columnwidth}{!}{
    \begin{tabular}{c|c c c}
        \Xhline{1pt}
        Dataset & Domains & Images & Classes \\
        \hline
        PACS~\cite{PACS} & $4$ & $9,991$ & $7$ \\
        VLCS~\cite{VLCS} & $4$ & $10,729$ & $5$ \\
        OfficeHome~\cite{OfficeHome} & $4$ & $15,588$ & $65$ \\
        TerraIncognita~\cite{TerraInc} & $4$ & $24,788$ & $10$ \\
        DomainNet~\cite{DomainNet} & $6$ & $586,575$ & $345$ \\
        \Xhline{1pt}
    \end{tabular}
    }
    \label{tab:dataset}
\end{table}

\subsection{Datasets and Implementation Details}
We evaluated our MeCAM against other competing methods on five public DG benchmarks, including PACS~\cite{PACS}, VLCS~\cite{VLCS}, OfficeHome~\cite{OfficeHome}, TerraIncognita~\cite{TerraInc}, and DomainNet~\cite{DomainNet}. The details of each dataset can be found in Table~\ref{tab:dataset}.

\begin{table*}[!t]
    \caption{Average accuracy and standard error ($mean_{\pm std}$) of our MeCAM and existing DG methods calculated across three trials on five public DG datasets. The best results are highlighted in \textbf{bold}. The results denoted by $\dag$ are taken from~\cite{SAGM} and~\cite{MADG}, while the results marked by $\ddag$ are inherited directly from the original source.}
    \centering
    \resizebox{0.91\textwidth}{!}{
    \begin{tabular}{c|c c c c c|c}
        \Xhline{1pt}
        Algorithm & 
        PACS & 
        VLCS & 
        OfficeHome & 
        TerraInc & 
        DomainNet &
        Average \\ 
        \hline
        MMD$^\dag$~\cite{MMD} & 
        $84.7_{\pm0.5}$ & 
        $77.5_{\pm0.9}$ & 
        $66.3_{\pm0.1}$ & 
        $42.2_{\pm1.6}$ & 
        $23.4_{\pm9.5}$ & 
        $58.8$ \\
        Mixstyle$^\dag$~\cite{MixStyle} & 
        $85.2_{\pm0.3}$ & 
        $77.9_{\pm0.5}$ & 
        $60.4_{\pm0.3}$ & 
        $44.0_{\pm0.7}$ & 
        $34.0_{\pm0.1}$ & 
        $60.3$ \\
        GroupDRO$^\dag$~\cite{GroupDRO} & 
        $84.4_{\pm0.8}$ & 
        $76.7_{\pm0.6}$ & 
        $66.0_{\pm0.7}$ & 
        $43.2_{\pm1.1}$ & 
        $33.3_{\pm0.2}$ & 
        $60.7$ \\
        IRM$^\dag$~\cite{IRM} & 
        $83.5_{\pm0.8}$ & 
        $78.5_{\pm0.5}$ & 
        $64.3_{\pm2.2}$ & 
        $47.6_{\pm0.8}$ & 
        $33.9_{\pm2.8}$ & 
        $61.6$ \\
        ARM$^\dag$~\cite{ARM} & 
        $85.1_{\pm0.4}$ & 
        $77.6_{\pm0.3}$ & 
        $64.8_{\pm0.3}$ & 
        $45.5_{\pm0.3}$ & 
        $35.5_{\pm0.2}$ & 
        $61.7$ \\
        VREx$^\dag$~\cite{VREx} & 
        $84.9_{\pm0.6}$ & 
        $78.3_{\pm0.2}$ & 
        $66.4_{\pm0.6}$ & 
        $46.4_{\pm0.6}$ & 
        $33.6_{\pm2.9}$ & 
        $61.9$ \\
        AND-mask$^\ddag$~\cite{SAND-mask} &
        $86.4_{\pm0.4}$ & 
        $76.4_{\pm0.4}$ & 
        $66.1_{\pm0.2}$ & 
        $49.8_{\pm0.4}$ & 
        $37.9_{\pm0.6}$ & 
        $63.3$ \\
        CDANN$^\dag$~\cite{CDANN} & 
        $82.6_{\pm0.9}$ & 
        $77.5_{\pm0.1}$ & 
        $65.8_{\pm1.3}$ &
        $45.8_{\pm1.6}$ & 
        $38.3_{\pm0.3}$ & 
        $62.0$ \\
        SAND-mask$^\ddag$~\cite{SAND-mask} & 
        $85.9_{\pm0.4}$ & 
        $76.2_{\pm0.5}$ & 
        $65.9_{\pm0.5}$ & 
        $50.2_{\pm0.1}$ & 
        $32.3_{\pm0.6}$ & 
        $62.1$ \\
        DANN$^\dag$~\cite{DANN} & 
        $83.6_{\pm0.4}$ & 
        $78.6_{\pm0.4}$ & 
        $65.9_{\pm0.6}$ & 
        $46.7_{\pm0.5}$ & 
        $38.3_{\pm0.1}$ & 
        $62.6$ \\
        MTL$^\dag$~\cite{MTL} & 
        $84.6_{\pm0.5}$ & 
        $77.2_{\pm0.4}$ & 
        $66.4_{\pm0.5}$ & 
        $45.6_{\pm1.2}$ & 
        $40.6_{\pm0.1}$ & 
        $62.9$ \\
        Mixup$^\dag$~\cite{Mixup} & 
        $84.6_{\pm0.6}$ & 
        $77.4_{\pm0.6}$ & 
        $68.1_{\pm0.3}$ & 
        $47.9_{\pm0.8}$ & 
        $39.2_{\pm0.1}$ & 
        $63.4$ \\
        MLDG$^\dag$~\cite{MLDG} & 
        $84.9_{\pm1.0}$ & 
        $77.2_{\pm0.4}$ & 
        $66.8_{\pm0.6}$ & 
        $47.7_{\pm0.9}$ & 
        $41.2_{\pm0.1}$ & 
        $63.6$ \\
        ERM$^\dag$~\cite{ERM} & 
        $85.5_{\pm0.2}$ & 
        $77.3_{\pm0.4}$ & 
        $66.5_{\pm0.3}$ & 
        $46.1_{\pm1.8}$ & 
        $43.8_{\pm0.1}$ & 
        $63.9$ \\
        Fish$^\ddag$~\cite{Fish} & 
        $85.5_{\pm0.3}$ & 
        $77.8_{\pm0.3}$ & 
        $68.6_{\pm0.4}$ & 
        $45.1_{\pm1.3}$ & 
        $42.7_{\pm0.2}$ & 
        $63.9$ \\        
        SagNet$^\dag$~\cite{SagNet} & 
        $86.3_{\pm0.2}$ & 
        $77.8_{\pm0.5}$ & 
        $68.1_{\pm0.1}$ & 
        $48.6_{\pm1.0}$ & 
        $40.3_{\pm0.1}$ & 
        $64.2$ \\
        SelfReg$^\ddag$~\cite{SelfReg} & 
        $85.6_{\pm0.4}$ & 
        $77.8_{\pm0.9}$ & 
        $67.9_{\pm0.7}$ & 
        $47.0_{\pm0.3}$ & 
        $42.8_{\pm0.0}$ & 
        $64.2$ \\
        mDSDI$^\ddag$~\cite{mDSDI} &
        $86.2_{\pm0.2}$ & 
        $79.0_{\pm0.3}$ & 
        $69.2_{\pm0.4}$ & 
        $48.1_{\pm1.4}$ & 
        $42.8_{\pm0.1}$ & 
        $65.1$ \\        
        Fishr$^\dag$~\cite{Fishr} &
        $86.9_{\pm0.2}$ & 
        $78.2_{\pm0.2}$ & 
        $68.2_{\pm0.2}$ & 
        $53.6_{\pm0.4}$ & 
        $41.8_{\pm0.2}$ & 
        $65.7$ \\
        MIRO$^\ddag$~\cite{MIRO} &
        $85.4_{\pm0.4}$ & 
        $79.0_{\pm0.0}$ & 
        $70.5_{\pm0.4}$ & 
        $50.4_{\pm1.1}$ & 
        $44.3_{\pm0.2}$ & 
        $65.9$ \\
        LP-FT$^\ddag$~\cite{LP-FT} &
        $84.6_{\pm0.8}$ & 
        $76.7_{\pm1.5}$ & 
        $65.0_{\pm0.2}$ & 
        $47.1_{\pm0.7}$ & 
        $43.0_{\pm0.1}$ & 
        $63.3$ \\
        MADG$^\dag$~\cite{MADG} &
        $86.5_{\pm0.4}$ & 
        $78.7_{\pm0.2}$ & 
        $\textbf{71.3}_{\pm0.3}$ & 
        $\textbf{53.7}_{\pm0.5}$ & 
        $39.9_{\pm0.4}$ & 
        $66.0$ \\
        \hline
        MeCAM (Ours) &
        $\textbf{88.3}_{\pm0.3}$ & 
        $\textbf{80.3}_{\pm0.3}$ & 
        $70.4_{\pm0.4}$ & 
        $48.9_{\pm1.1}$ & 
        $\textbf{45.0}_{\pm0.0}$ & 
        $\textbf{66.6}$ \\ 
        \Xhline{1pt}
    \end{tabular}
    }
    \label{tab:comparison_DG}
\end{table*}

For a fair comparison, we adopted the training and evaluation protocols from DomainBed~\cite{gulrajanisearch}, where hyperparameter tuning was performed without access to test data. All models were trained on DomainNet for 15,000 iterations and on the other datasets for 5,000 iterations, unless otherwise specified. We used the leave-one-domain-out protocol for evaluation, where one domain is held out as the target (test) domain, and the remaining domains serve as the source (training) domains.
For each dataset, $20\%$ of the training data was used for validation and model selection. We utilized the average accuracy across all domains to measure the performance of each model and performed three experimental trials (\emph{i.e.}, setting the random seed to 0, 1, and 2) to compute the mean value and standard error of the performance metric.
The default backbone used in this study is the ResNet-50~\cite{ResNet} pretrained on ImageNet~\cite{ImageNet}. We utilized the Adam optimizer~\cite{Adam} as the base optimizer for MeCAM. The optimal hyperparameter settings of MeCAM for each dataset are provided in Supplementary Material B.

\subsection{Comparison with Existing DG Methods}
We compared our MeCAM with 22 existing conventional DG methods~\cite{MMD,MixStyle,GroupDRO,IRM,ARM,VREx,CDANN,SAND-mask,DANN,MTL,Mixup,MLDG,ERM,Fish,SagNet,SelfReg,mDSDI,Fishr,MIRO,LP-FT,MADG} to evaluate its generalization ability. The results were presented in Table~\ref{tab:comparison_DG}. 
Compared to the ERM algorithm~\cite{ERM}, which can be treated as the baseline, MeCAM consistently demonstrates a substantial improvement across all benchmarks, with an average performance gain of $2.7\%$.
MeCAM outperforms most methods on all datasets, with particularly notable improvements on PACS, VLCS, and DomainNet. Specifically, MeCAM achieves a $1.8\%$ improvement over MADG on PACS, a $1.3\%$ improvement over MIRO on VLCS, and a $0.7\%$ improvement over MIRO on DomainNet.
Although MADG achieves the highest performance on OfficeHome and TerraIncognita, its performance on DomainNet is relatively weak. DomainNet is the largest dataset with over half a million images across 345 categories and collected from six distinct domains, posing a significant challenge. Our MeCAM surpasses other DG methods on this dataset, highlighting its robustness in handling such a large-scale and complex dataset. Overall, these results demonstrate that MeCAM improves the model generalization across a wide range of DG tasks.

\subsection{Comparison with Optimizers and Sharpness-Based DG Methods}
Since MeCAM can be viewed as an optimizer designed to improve the update of model parameter, we further compared it with six other optimizers and six sharpness-based DG methods from the perspective of model optimization. The results were shown in Table~\ref{tab:comparison_Opt}. 
It reveals that:
(1) AdaHessian achieves the best overall performance among the optimizers, indicating the effectiveness of Hessian-based optimization for enhancing generalization;
(2) even the sharpness-based method with the lowest performance, \emph{i.e.}, SAM, outperforms all the optimizers, demonstrating the importance of reducing sharpness of the loss landscape for better generalization; 
(3) different datasets exhibit preferences for different algorithms, \emph{e.g.}, AdaHessian outperforms AdamW on PACS, VLCS, and DomainNet, but possesses lower performance on OfficeHome and DomainNet; 
and
(4) MeCAM consistently outperforms other optimizers and sharpness-based methods across all benchmark datasets, confirming that simultaneously minimizing the surrogate gap of SAM and meta-learning for reducing curvature around the local minima leads to superior generalization.

\begin{table*}[!t]
    \caption{Average accuracy and standard error ($mean_{\pm std}$) of our MeCAM, optimizers, and existing sharpness-based DG methods calculated across three trials on five DG datasets. The best results are highlighted in \textbf{bold}. Results marked with $\dag$ and * are inherited from~\cite{FAD} and obtained by reproduction, respectively.}
    \centering
    \resizebox{\textwidth}{!}{
    \begin{tabular}{c|c|c c c c c|c}
        \Xhline{1pt}
        \multicolumn{2}{c|}{Algorithm} & 
        PACS & 
        VLCS & 
        OfficeHome & 
        TerraInc & 
        DomainNet &
        Average \\ 
        \hline
        {\multirow{6}*{Optimizer}}
        & Adam$^\dag$~\cite{Adam}	&
        $84.2_{\pm0.6}$ & 
        $77.3_{\pm1.3}$ & 
        $67.6_{\pm0.4}$ & 
        $44.4_{\pm0.8}$ & 
        $43.0_{\pm0.1}$ & 
        $63.3$ \\ 	
        & AdamW$^\dag$~\cite{AdamW} &
        $83.6_{\pm1.5}$ & 
        $77.4_{\pm0.8}$ & 
        $68.8_{\pm0.6}$ & 
        $45.2_{\pm1.4}$ & 
        $43.4_{\pm0.1}$ & 
        $63.7$ \\ 
        & SGD$^\dag$~\cite{SGD}	&
        $79.9_{\pm1.4}$ & 
        $78.1_{\pm0.2}$ & 
        $68.5_{\pm0.3}$ & 
        $44.9_{\pm1.8}$ & 
        $43.2_{\pm0.1}$ & 
        $62.9$ \\ 	 
        & YOGI$^\dag$~\cite{YOGI}	&
        $81.2_{\pm0.4}$ & 
        $77.6_{\pm0.6}$ & 
        $68.3_{\pm0.3}$ & 
        $45.4_{\pm0.5}$ & 
        $43.5_{\pm0.0}$ & 
        $63.2$ \\  
        & AdaBelief$^\dag$~\cite{AdaBelief}	&
        $84.6_{\pm0.6}$ & 
        $78.4_{\pm0.4}$ & 
        $68.0_{\pm0.9}$ & 
        $45.2_{\pm2.0}$ & 
        $43.5_{\pm0.1}$ & 
        $63.9$ \\ 
        & AdaHessian$^\dag$~\cite{AdaHessian}	&
        $84.5_{\pm1.0}$ & 
        $78.6_{\pm0.8}$ & 
        $68.4_{\pm0.9}$ & 
        $44.4_{\pm0.5}$ & 
        $44.4_{\pm0.1}$ & 
        $64.1$ \\ 	
        \hline
        {\multirow{6}*{Sharpness-based}}
        & SAM$^\dag$~\cite{SAM}	&
        $85.3_{\pm1.0}$ & 
        $78.2_{\pm0.5}$ & 
        $68.0_{\pm0.8}$ & 
        $45.7_{\pm0.9}$ & 
        $43.4_{\pm0.1}$ & 
        $64.1$ \\ 	
        & GAM$^\dag$~\cite{GAM}	&
        $86.1_{\pm0.6}$ & 
        $78.5_{\pm0.4}$ & 
        $68.2_{\pm1.0}$ & 
        $45.2_{\pm0.6}$ & 
        $43.8_{\pm0.1}$ & 
        $64.4$ \\ 	
        & SAGM*~\cite{SAGM}	&
        $86.9_{\pm0.3}$ & 
        $79.1_{\pm1.0}$ & 
        $69.4_{\pm0.1}$ & 
        $48.6_{\pm1.5}$ & 
        $44.7_{\pm0.2}$ & 
        $65.7$ \\ 	
        & FAD$^\dag$~\cite{FAD}	&
        $88.2_{\pm0.5}$ & 
        $78.9_{\pm0.8}$ & 
        $69.2_{\pm0.5}$ & 
        $45.7_{\pm1.0}$ & 
        $44.4_{\pm0.1}$ & 
        $65.3$ \\ 	
        & CRSAM*~\cite{CRSAM}	&
        $85.4_{\pm0.7}$ & 
        $79.1_{\pm0.6}$ & 
        $68.9_{\pm1.0}$ & 
        $45.3_{\pm0.4}$ & 
        $44.3_{\pm0.1}$ & 
        $64.6$ \\ 	
        & FSAM*~\cite{FSAM}	&
        $86.5_{\pm0.3}$ & 
        $79.4_{\pm0.1}$ & 
        $70.2_{\pm0.1}$ & 
        $46.1_{\pm0.6}$ & 
        $44.9_{\pm0.1}$ & 
        $65.4$ \\ 	
        \hline
        \multicolumn{2}{c|}{MeCAM (Ours)} &
        $\textbf{88.3}_{\pm0.3}$ & 
        $\textbf{80.3}_{\pm0.3}$ & 
        $\textbf{70.4}_{\pm0.4}$ & 
        $\textbf{48.9}_{\pm1.1}$ & 
        $\textbf{45.0}_{\pm0.0}$ & 
        $\textbf{66.6}$ \\ 
        \Xhline{1pt}
    \end{tabular}
    }
    \label{tab:comparison_Opt}
\end{table*}

\begin{table*}[!t]
    \caption{Performance of our MeCAM, three optimizers, and FAD integrated with other DG methods on five public DG datasets. The best results are highlighted in \textbf{bold}. The results denoted by $\dag$ are inherited from~\cite{FAD}.}
    \centering
    \resizebox{0.9\textwidth}{!}{
    \begin{tabular}{c|c c c c c|c}
        \Xhline{1pt}
        Algorithm & 
        PACS & 
        VLCS & 
        OfficeHome & 
        TerraInc & 
        DomainNet &
        Average \\ 
        \hline
        Adam + SWAD$^\dag$ & 
        $86.8$ & 
        $79.1$ & 
        $70.1$ & 
        $46.5$ & 
        $44.1$ & 
        $65.3$ \\ 
        AdamW + SWAD$^\dag$ & 
        $87.0$ & 
        $78.5$ & 
        $70.8$ & 
        $46.9$ & 
        $45.0$ & 
        $65.6$ \\
        SGD + SWAD$^\dag$ & 
        $85.2$ & 
        $79.1$ & 
        $71.0$ & 
        $46.7$ & 
        $42.8$ & 
        $65.0$ \\
        FAD + SWAD$^\dag$ & 
        $88.5$ & 
        $\textbf{79.8}$ & 
        $\textbf{71.8}$ & 
        $47.5$ & 
        $45.0$ & 
        $66.5$ \\
        MeCAM (Ours) + SWAD &
        $\textbf{88.9}$ & 
        $\textbf{79.8}$ & 
        $70.8$ & 
        $\textbf{48.9}$ & 
        $\textbf{45.9}$ & 
        $\textbf{66.9}$ \\ 
        \hline
        Adam + CORAL$^\dag$ & 
        $86.0$ & 
        $78.9$ & 
        $68.7$ & 
        $43.7$ & 
        $44.5$ & 
        $64.5$ \\ 
        AdamW + CORAL$^\dag$ & 
        $86.4$ & 
        $79.5$ & 
        $69.8$ & 
        $45.0$ & 
        $44.9$ & 
        $65.1$ \\ 
        SGD + CORAL$^\dag$ & 
        $85.6$ & 
        $78.2$ & 
        $69.5$ & 
        $45.8$ & 
        $44.6$ & 
        $64.7$ \\ 
        FAD + CORAL$^\dag$ & 
        $\textbf{88.5}$ & 
        $78.9$ & 
        $70.8$ & 
        $46.1$ & 
        $44.9$ & 
        $65.9$ \\ 
        MeCAM (Ours) + CORAL &
        $88.0$ & 
        $\textbf{79.7}$ & 
        $\textbf{71.0}$ & 
        $\textbf{49.9}$ & 
        $\textbf{45.0}$ & 
        $\textbf{66.7}$ \\
        \hline
        Adam + RSC$^\dag$ & 
        $84.5$ & 
        $77.9$ & 
        $65.7$ & 
        $44.5$ & 
        $42.8$ & 
        $63.1$ \\ 
        AdamW + RSC$^\dag$ & 
        $83.4$ & 
        $77.5$ & 
        $66.3$ & 
        $45.1$ & 
        $42.4$ & 
        $62.9$ \\ 
        SGD + RSC$^\dag$ & 
        $82.6$ & 
        $78.1$ & 
        $67.0$ & 
        $43.9$ & 
        $43.5$ & 
        $63.0$ \\
        FAD + RSC$^\dag$ & 
        $86.9$ & 
        $77.6$ & 
        $68.6$ & 
        $46.2$ & 
        $44.1$ & 
        $64.7$ \\
        MeCAM (Ours) + RSC &
        $\textbf{87.2}$ & 
        $\textbf{79.8}$ & 
        $\textbf{70.2}$ & 
        $\textbf{47.3}$ & 
        $\textbf{44.7}$ & 
        $\textbf{65.8}$ \\ 
        \Xhline{1pt}
    \end{tabular}
    }
    \label{tab:extensibility}
\end{table*}

\subsection{Extensibility Analysis}
To evaluate the extensibility of our MeCAM, we integrated it with three representative DG methods (\emph{i.e.}, SWAD~\cite{SWAD}, CORAL~\cite{CORAL}, and RSC~\cite{RSC}), and repeated experiments on five DG datasets.
The optimal hyperparameter configurations listed in Supplementary Material B were directly applied to MeCAM during its integration with other DG methods.
Therefore, no additional hyperparameter search was performed.
The results were presented in Table~\ref{tab:extensibility}.
It shows that the combinations involved our MeCAM consistently outperforms others across most scenarios and also achieve the best overall performance, demonstrating the superior extensibility of our MeCAM.

\begin{figure*}[!ht]
  \centering
  \includegraphics[width=\textwidth]{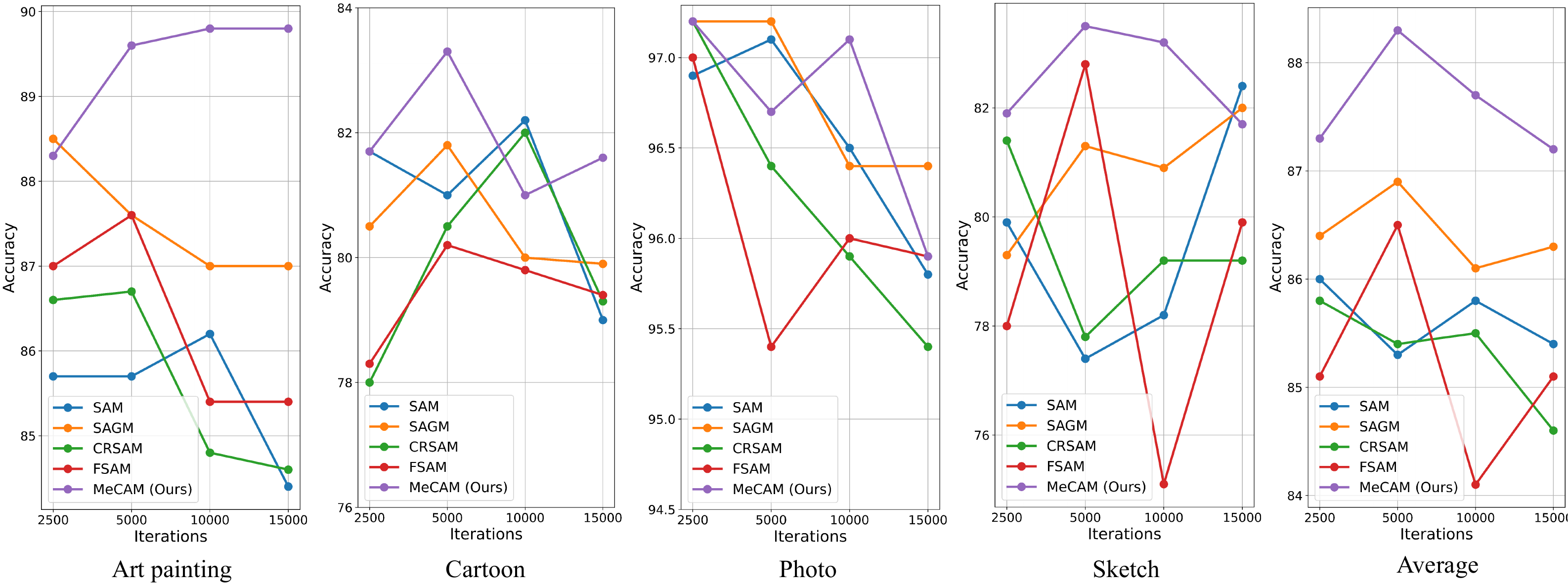}
   \caption{Accuracy of our MeCAM and other sharpness-based methods across various training iterations using each domain of PACS as the target domain. For each figure, the X-axis denotes the training iterations, where $5,000$ is the default configuration, and the Y-axis indicates the accuracy.}
\label{fig:iterations}
\end{figure*}

\subsection{Evaluation on Various Training Iterations}
To explore how the training iteration takes effect on our MeCAM and other sharpness-based methods, we further evaluated them on various training iterations using each domain of PACS as the target domain. The results were displayed in Figure~\ref{fig:iterations}.
It shows that MeCAM outperforms other competing methods in most scenarios, particularly in the target domains of ``Art painting'', ``Cartoon'', and ``Sketch''. 
On the "Photo" domain, it is hard to distinguish which method is superior over others, as performing classification on this domain is relatively simple, where all methods exhibit an accuracy over $95\%$.
Additionally, MeCAM achieves the best overall performance across all training iterations, demonstrating its superior generalization ability and robustness.

\begin{table}[!ht]
    \caption{Curvature metric of our MeCAM and five competing methods on the PACS dataset with various values of $\rho$. The best result of each column is highlighted in \textbf{bold}.}
    \centering
    \resizebox{1\columnwidth}{!}{
    \begin{tabular}{c|c c c c c}
        \Xhline{1pt}
        Algorithm & 
        $\rho=0.01$ & 
        $\rho=0.05$ & 
        $\rho=0.1$ & 
        $\rho=0.2$ & 
        $\rho=0.5$ \\ 
        \hline
        ERM~\cite{ERM} & 
        $4.4e^{-3}$ & 
        $8.6e^{-2}$ & 
        $2.7e^{-1}$ & 
        $2.0$ & 
        $33.5$ \\
        SAM~\cite{SAM} & 
        $5.5e^{-4}$ & 
        $1.4e^{-2}$ & 
        $5.5e^{-2}$ & 
        $2.6e^{-1}$ & 
        $8.6$ \\
        SAGM~\cite{SAGM} & 
        $1.0e^{-3}$ & 
        $3.4e^{-2}$ & 
        $1.4e^{-1}$ & 
        $9.0e^{-1}$ & 
        $15.5$ \\
        CRSAM~\cite{CRSAM} & 
        $3.6e^{-4}$ & 
        $1.3e^{-2}$ & 
        $5.4e^{-2}$ & 
        $2.6e^{-1}$ & 
        $6.0$ \\       
        FSAM~\cite{FSAM} & 
        $2.4e^{-3}$ & 
        $4.2e^{-2}$ & 
        $1.1e^{-1}$ & 
        $4.2e^{-1}$ & 
        $5.0$ \\       
        \hline
        MeCAM (Ours) & 
        $\textbf{3.0e}^{\textbf{-4}}$ & 
        $\textbf{7.5e}^{\textbf{-3}}$ & 
        $\textbf{2.9e}^{\textbf{-2}}$ & 
        $\textbf{1.2e}^{\textbf{-1}}$ & 
        $\textbf{2.5}$ \\         
        \Xhline{1pt}
    \end{tabular}
    }
    \label{tab:curvature}
\end{table}

\subsection{Curvature Comparison on PACS}
In this experiment, we evaluated the curvature metric $\mathcal{C}$ around the local minima obtained by our MeCAM and other competing methods. 
Based on the above approximation, we utilized various values of $\rho$ to compute $\mathcal{C}$ on the PACS dataset by
$\mathcal{C}(f(\theta))=\alpha\frac{| f(\theta+\delta)+f(\theta-\delta)-2f(\theta) |}{\|\nabla f(\theta)\|^2+1}$, where $\alpha$ is omitted for convenient comparison. The results were listed in Table~\ref{tab:curvature}, with smaller curvature indicating superior algorithm.
It can be found that our MeCAM consistently outperforms other competing methods across all values of $\rho$ and improves $\mathcal{C}$ significantly as $\rho$ increases, demonstrating the robustness of our MeCAM against large perturbations.

\begin{figure}[!ht]
  \centering
  \includegraphics[width=\columnwidth]{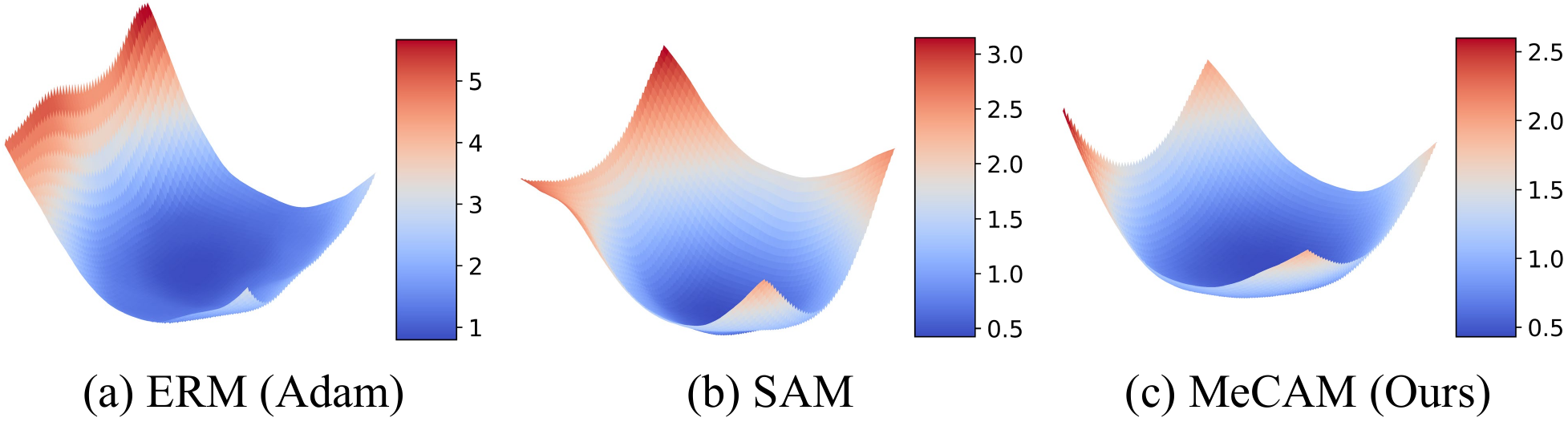}
   \caption{Visualization of the loss landscapes of ERM, SAM, and our MeCAM on the PACS dataset. The number represents the loss value.}
\label{fig:LossSurface}
\end{figure}

\subsection{Visualization of Loss Landscapes}
To demonstrate that MeCAM achieves a flatter minima, we visualized the loss landscapes of the ResNet50 trained using ERM (Adam), SAM, and our MeCAM on the PACS dataset with 5,000 iterations. 
We utilized the visualization techniques proposed in~\cite{li2018visualizing} and plotted the loss values along two randomly sampled orthogonal Gaussian perturbations around the local minima. 
As shown in Figure~\ref{fig:LossSurface}, the results reveal that (1) ERM is very sensitive to perturbations; (2) our MeCAM converges to a flatter minima compared to both ERM and SAM; and (3) our MeCAM also results in a flat minima with lower loss values.
These results demonstrate the superiority of MeCAM in decreasing the sharpness around the local minima.

\section{Conclusion}
\label{sec:conclu}
In this paper, to tackle the limitations in SAM and its existing variants, we introduce a more rational training process for sharpness-based methods, incorporating a curvature metric that can better measure the sharpness. This metric focuses on the curvature around the local minima and can bound the maximum eigenvalue of the Hessian matrix. 
We propose a novel generalization approach, MeCAM, which minimizes both the surrogate gap of sharpness-aware minimization and the surrogate gap of meta-learning to seek a flatter minima.
We demonstrate that MeCAM effectively controls the generalization error and achieves a competitive convergence rate. 
Compared to existing conventional DG methods and sharpness-based methods, MeCAM shows superior generalization performance across various DG datasets.
In our future work, we plan to improve MeCAM by: (1) developing better methods for computing the step-size in the central difference approximation; and (2) exploring the combination of MeCAM with other meta-learning paradigms or investigating other ways to synthesize the virtual meta-test domain.

\newpage

{
    \small
    \bibliographystyle{ieeenat_fullname}
    \bibliography{references.bib}
}

\clearpage
\setcounter{page}{1}
\maketitlesupplementary
\appendix

\renewcommand{\thetable}{\Roman{table}}
\renewcommand{\thefigure}{\Roman{figure}}
\setcounter{table}{0}
\setcounter{figure}{0}

In the supplementary material, we provide more details about the proofs omitted in the manuscript (Section~\ref{Proof}), the optimal hyperparameter configurations for MeCAM on each dataset (Section~\ref{Hyperparameter}), the full results of Table 2 and Table 3 in the manuscript (Section~\ref{Full_Result}), and additional experiments (Section~\ref{Experiment}).

\section{Proofs}
\label{Proof}

\subsection{Proof of Proposition 3.3} 
\textit{Proof.} Suppose the training set has $n$ elements drawn i.i.d. from the true distribution, and denote the loss on the training set as $\hat{f}(\theta)$.
Let $\theta$ be learned from the training set with a number of $k$. Then following the proof of Theorem 2 in~\cite{SAM}, with probability at least $1-\zeta$, we have:
\begin{equation}
\begin{aligned}
&\mathbb{E}_{\delta_i \sim \mathcal{N}(0,\sigma^2)}[f(\theta+\delta)] \le \mathbb{E}_{\delta_i \sim \mathcal{N}(0,\sigma^2)}[\hat{f}(\theta+\delta)] + \\ 
&\sqrt{\frac{\frac{1}{4}k\log (1+\frac{\| \theta \|^2}{k\sigma^2})+\frac{1}{4}+\log \frac{n}{\zeta}+2\log(6n+3k)}{n-1}}.
\end{aligned}
\label{dt_1}
\end{equation}
Since $\delta_i \sim \mathcal{N}(0,\sigma^2)$, $\frac{\| \delta \|^2}{\sigma^2}$ has a chi-square distribution. By Lemma 1 in~\cite{laurent2000adaptive}, we have that for any $t>0$:
\begin{equation}
\begin{aligned}
P(\frac{\| \delta \|^2}{\sigma^2} - k \ge 2 \sqrt{kt} + 2t) \le \exp(-t).
\end{aligned}
\label{PAC_Bayesian}
\end{equation}
By $t=\ln\sqrt{n}$, with probability at least $1-\frac{1}{\sqrt{n}}$, we have:
\begin{equation}
\begin{aligned}
\| \delta \|^2 &\le \sigma^2(k + 2 \sqrt{k\ln \sqrt{n}} + 2 \ln \sqrt{n}) \\
&\le \sigma^2(\sqrt{k}+\sqrt{\ln n})^2 \le \rho^2.
\end{aligned}
\label{delta_rho}
\end{equation}
According to the Taylor expansion, when $\nabla \hat{f}(\theta)$ is approaching to the local minimum and $\delta$ is small, we can obtain:
\begin{equation}
\begin{aligned}
\hat{f}(\theta+\delta) &\approx \hat{f}(\theta) + \delta \nabla \hat{f}(\theta) + \frac{\delta^2}{2} \hat{H}(\theta)\\
&\le \hat{f}(\theta) + \hat{H}(\theta) \\
&\approx \hat{f}(\theta) + \alpha(\hat{f}(\theta+\delta)+\hat{f}(\theta-\delta)-2\hat{f}(\theta))
\end{aligned}
\label{sam_ours}
\end{equation}
where $\hat{H}(\theta)$ is the Hessian obtained on the training set.
Following the proof of proposition 4.3 in~\cite{GAM} and combining Eq. (\ref{sam_ours}), Eq. (\ref{delta_rho}), and Eq. (\ref{PAC_Bayesian}), we have:
\begin{equation}
\begin{aligned}
&\mathbb{E}_{\delta_i \sim \mathcal{N}(0,\sigma^2)}[f(\theta+\delta)] \le
\hat{f}(\theta) + \\
&\alpha(\hat{f}(\theta+\delta)+\hat{f}(\theta-\delta)-2\hat{f}(\theta)) + \frac{M}{\sqrt{n}} + \\
&\sqrt{\frac{\frac{1}{4}k\log (1+\frac{\| \theta \|^2 (1+\sqrt{\frac{\ln n}{k}})^2}{\rho^2})+\frac{1}{4}+\log \frac{n}{\zeta}+2\log(6n+3k)}{n-1}},
\end{aligned}
\end{equation}
where the loss function calculated on each data is bounded by $M$.

\subsection{Proof of Theorem 3.4} 
\textit{Proof.} For simplicity, we utilize the SGD optimizer as the base optimizer to analyze the convergence rate of our CAM.
For other optimizers, such as Adam, similar results can be derived by extending our proof.
Since there are three items in $F(\theta)$, we first analyze each one individually and then combine the results together.

\subsubsection{Convergence W.R.T. Function $f(\theta)$}
\label{convergence_oracle}
For simplicity, the update $d_t$ of the model parameter $\theta$ at step $t$ can be written as: 
\begin{equation}
\begin{aligned}
d_t = -\eta_t(\gamma g_t+\alpha g_t^{sam}+\alpha g_t^{meta}), \gamma = 1 - 2\alpha.
\end{aligned}
\label{dt_1}
\end{equation}
Suppose $f$ is $L$-Lipschitz smooth. By the definitions of $d_t=\theta_{t+1}-\theta_t$, $f(\theta_t^{sam}) = f(\theta_t + \delta_t)$, and $f(\theta_t^{meta}) = f(\theta_t - \delta_t)$, we have:
\begin{equation}
\begin{aligned}
f(\theta_{t+1}) &\le f(\theta_t) + \langle \nabla f(\theta_t), \theta_{t+1}-\theta_t \rangle +\frac{L}{2}\| \theta_{t+1}-\theta_t \|^2 \\
&=f(\theta_t) + \langle \nabla f(\theta_t), d_t \rangle +\frac{L}{2}\| d_t \|^2 \\
&=f(\theta_t) + \langle \nabla f(\theta_t), -\eta_t(\gamma g_t+\alpha g_t^{sam}+\alpha g_t^{meta}) \rangle \\
&+\frac{L}{2}\| d_t \|^2.
\end{aligned}
\label{Lipschitz_1}
\end{equation}
By the assumptions of $\mathbb{E}[g_t] = \nabla f(\theta_t)$, $\mathbb{E}[g_t^{sam}] = \nabla f(\theta_t^{sam})$, and $\mathbb{E}[g_t^{meta}] = \nabla f(\theta_t^{meta})$, we take the expectation conditioned on the observations up to step $t$ for both sides and then obtain:
\begin{equation}
\begin{aligned}
\mathbb{E}[f(\theta_{t+1})] &\le f(\theta_{t}) + \langle \nabla f(\theta_t), -\eta_t(\gamma \mathbb{E}[g_t] + \alpha \mathbb{E}[g_t^{sam}] \\
& + \alpha \mathbb{E}[g_t^{meta}]) \rangle 
+\frac{L}{2}\mathbb{E}[\| d_t \|^2] \\
&=f(\theta_{t}) - \eta_t (\gamma \| \nabla f(\theta_t) \|^2 \\
&+ \alpha \langle \nabla f(\theta_t), \nabla f(\theta_t^{sam}) \rangle \\
&+ \alpha \langle \nabla f(\theta_t), \nabla f(\theta_t^{meta}) \rangle) + \frac{L}{2}\mathbb{E}[\| d_t \|^2].
\end{aligned}
\label{Exp_1_1}
\end{equation}
Since perturbation $\delta_t$ is small, by Taylor expansion, $\nabla f(\theta_t^{sam})$ and $\nabla f(\theta_t^{meta})$ can be approximated as:
\begin{equation}
\begin{aligned}
\nabla f(\theta_t^{sam}) = \nabla f(\theta_t + \delta_t) \approx \nabla f(\theta_t) + \delta_t H(\theta_t),
\end{aligned}
\label{Taylor_sam_1}
\end{equation}
and
\begin{equation}
\begin{aligned}
\nabla f(\theta_t^{meta}) = \nabla f(\theta_t - \delta_t) \approx \nabla f(\theta_t) - \delta_t H(\theta_t),
\end{aligned}
\label{Taylor_meta_1}
\end{equation}
where $H(\theta_t)$ is the Hessian and the Taylor remainder is omitted.
Plug Eq. (\ref{Taylor_sam_1}) and Eq. (\ref{Taylor_meta_1}) into Eq. (\ref{Exp_1_1}), with the definition of $\gamma = 1 - 2\alpha$, we have:
\begin{equation}
\begin{aligned}
\mathbb{E}[f(\theta_{t+1})] &\le f(\theta_{t}) - \eta_t (\gamma + 2\alpha) \| \nabla f(\theta_t) \|^2 \\
&- \eta_t \alpha \delta_t \langle \nabla f(\theta_t), H(\theta_t) \rangle \\
&+ \eta_t \alpha \delta_t \langle \nabla f(\theta_t), H(\theta_t) \rangle + \frac{L}{2}\mathbb{E}[\| d_t \|^2] \\
&= f(\theta_{t}) - \eta_t \| \nabla f(\theta_t) \|^2 + \frac{L}{2}\mathbb{E}[\| d_t \|^2].
\end{aligned}
\label{Exp_1_2}
\end{equation}
Assume gradient is upper-bounded by $G$, Eq. (\ref{Exp_1_2}) can be rewritten as:
\begin{equation}
\begin{aligned}
\mathbb{E}[f(\theta_{t+1})] &\le f(\theta_{t}) - \eta_t \| \nabla f(\theta_t) \|^2 + \frac{L}{2}\mathbb{E}[G^2 \eta_t^2] \\
&= f(\theta_{t}) - \eta_t \| \nabla f(\theta_t) \|^2 + \frac{L}{2} G^2 \eta_t^2.
\end{aligned}
\label{Exp_1_3}
\end{equation}
By re-arranging above formula, we have:
\begin{equation}
\begin{aligned}
\eta_t \| \nabla f(\theta_t) \|^2 \le f(\theta_{t}) - \mathbb{E}[f(\theta_{t+1})] + \frac{L}{2} G^2 \eta_t^2.
\end{aligned}
\label{Exp_1_4}
\end{equation}
Perform telescope sum and take the expectation on each step $t$, we can obtain:
\begin{equation}
\begin{aligned}
\sum_{t=1}^T \eta_t \| \nabla f(\theta_t) \|^2 \le f(\theta_1) - \mathbb{E}[f(\theta_{T})] + \frac{L}{2} G^2 \sum_{t=1}^T \eta_t^2.
\end{aligned}
\label{Telescope_1_1}
\end{equation}
By setting $\eta_t = \frac{\eta_0}{\sqrt{t}}$, we have:
\begin{equation}
\begin{aligned}
\frac{\eta_0}{\sqrt{t}} \sum_{t=1}^T \| \nabla f(\theta_t) \|^2 &\le f(\theta_1) - f_{min} + \frac{L}{2} G^2 \eta_0^2 \sum_{t=1}^T \frac{1}{t} \\
&\le f(\theta_1) - f_{min} + \frac{L}{2} G^2 \eta_0^2 (1 + \log T).
\end{aligned}
\label{Telescope_1_2}
\end{equation}
Hence
\begin{equation}
\begin{aligned}
\frac{1}{T} \sum_{t=1}^T \| \nabla f(\theta_t) \|^2 \le 
\frac{C_1 + C_2 \log T}{\sqrt{T}}
\end{aligned}
\label{Convergence_1}
\end{equation}
implies the convergence rate w.r.t. $f(\theta)$ is $O(\frac{\log T}{\sqrt{T}})$, where $C_1$ and $C_2$ are some constants.

\subsubsection{Convergence W.R.T. Function $f(\theta^{sam})$}
\label{convergence_sam}
Denote the update $d_t$ of the model parameter $\theta$ at step $t$ as: 
\begin{equation}
\begin{aligned}
d_t = -\eta_t(\gamma g_t+\alpha g_t^{sam}+\alpha g_t^{meta}), \gamma = 1 - 2\alpha.
\end{aligned}
\label{dt_2}
\end{equation}
Define $d_t=\theta_{t+1}-\theta_t$, $f(\theta_t^{sam}) = f(\theta_t + \delta_t)$, and $f(\theta_t^{meta}) = f(\theta_t - \delta_t)$.
By $L$-smoothness of $f$, we have:
\begin{equation}
\begin{aligned}
f(\theta_{t+1}^{sam}) &\le f(\theta_t^{sam}) + \langle \nabla f(\theta_t^{sam}), \theta_{t+1}^{sam}-\theta_t^{sam} \rangle \\
&+\frac{L}{2}\| \theta_{t+1}^{sam}-\theta_t^{sam} \|^2 \\
&=f(\theta_t^{sam}) + \langle \nabla f(\theta_t^{sam}), \theta_{t+1}+\delta_{t+1}-\theta_t-\delta_t \rangle \\
&+\frac{L}{2}\| \theta_{t+1}+\delta_{t+1}-\theta_t-\delta_t \|^2.
\end{aligned}
\label{Lipschitz_2_1}
\end{equation}
According to the Cauchy–Schwarz inequality and AM-GM inequality, we re-arrange the above formula and then obtain:
\begin{equation}
\begin{aligned}
& f(\theta_{t+1}^{sam}) - f(\theta_t^{sam}) - \langle \nabla f(\theta_t^{sam}), d_t \rangle - L \| d_t \|^2 \le \\
& \langle \nabla f(\theta_t^{sam}), \delta_{t+1}-\delta_t \rangle + L \| \delta_{t+1}-\delta_t \|^2.
\end{aligned}
\label{Lipschitz_2_2}
\end{equation}
We first take the expectation conditioned on the observations up to step $t$ for the $LHS$. By reusing results from Eq. (\ref{Taylor_sam_1}) and Eq. (\ref{Taylor_meta_1}), we have:
\begin{equation}
\begin{aligned}
LHS = &\mathbb{E}[f(\theta_{t+1}^{sam})] - f(\theta_t^{sam}) - L\mathbb{E}[\| d_t \|^2] - \\
&\langle \nabla f(\theta_t^{sam}), -\eta_t(\gamma \mathbb{E}[g_t] + \alpha \mathbb{E}[g_t^{sam}] + \alpha \mathbb{E}[g_t^{meta}]) \rangle \\
= &\mathbb{E}[f(\theta_{t+1}^{sam})] - f(\theta_t^{sam}) - L\mathbb{E}[\| d_t \|^2] + \\
&\eta_t (\alpha \| \nabla f(\theta_t^{sam}) \|^2 + \alpha \langle \nabla f(\theta_t^{sam}), \nabla f(\theta_t^{meta}) \rangle + \\
&\gamma \langle \nabla f(\theta_t^{sam}), \nabla f(\theta_t) \rangle) \\
\approx & \mathbb{E}[f(\theta_{t+1}^{sam})] - f(\theta_t^{sam}) - L\mathbb{E}[\| d_t \|^2] + \\
& \eta_t(\alpha \| \nabla f(\theta_t^{sam}) \|^2 + \alpha \langle \nabla f(\theta_t^{sam}), \nabla f(\theta_t^{sam}) - \\
&2\delta_t H(\theta_t) \rangle + \gamma \langle \nabla f(\theta_t^{sam}), \nabla f(\theta_t^{sam}) - \delta_t H(\theta_t) \rangle ) \\
= & \mathbb{E}[f(\theta_{t+1}^{sam})] - f(\theta_t^{sam}) - L\mathbb{E}[\| d_t \|^2] + \\
&\eta_t (\| \nabla f(\theta_t^{sam}) \|^2 - \langle \nabla f(\theta_t^{sam}), \delta_t H(\theta_t) \rangle) \\
\ge &\mathbb{E}[f(\theta_{t+1}^{sam})] - f(\theta_t^{sam}) + \eta_t\| \nabla f(\theta_t^{sam}) \|^2 - \\
& L G \rho_t \eta_t - L G^2 \eta_t^2,
\end{aligned}
\label{Exp_2_1}
\end{equation}
where the last inequality is due to (1) max eigenvalue of $H$ is upper-bounded by $L$, (2) gradient is assumed to be upper-bounded by $G$, and (3) the definition of $\delta_t$ that $\delta_t = \rho_t \frac{g_t}{\|g_t\| + \epsilon}$ and $\| \rho_t \frac{g_t}{\|g_t\| + \epsilon} \| < 1$.

Inspired by~\cite{GSAM}, we take the expectation conditioned on the observations up to step $t$ for the $RHS$ of Eq. (\ref{Lipschitz_2_2}) and then obtain:
\begin{equation}
\begin{aligned}
RHS =& \langle \nabla f(\theta_t^{sam}), \mathbb{E}[\delta_{t+1} - \delta_t] \rangle + L \mathbb{E}[\| \delta_{t+1} - \delta_t \|^2] \\
\le& G \rho_t \mathbb{E}[\| \frac{g_{t+1}}{g_{t+1} + \epsilon} - \frac{g_t}{g_t + \epsilon} \|] + \\
& L\rho_t^2 \mathbb{E}[\| \frac{g_{t+1}}{g_{t+1} + \epsilon} - \frac{g_t}{g_t + \epsilon} \|^2]\\
\le& G \rho_t \phi_t + L \rho_t^2 \phi_t^2,
\end{aligned}
\label{Exp_2_2}
\end{equation}
where $\phi_t$ denotes the angle between $\nabla f(\theta_{t+1})$ and $\nabla f(\theta_t)$.
According to the analysis in~\cite{GSAM}, when $\eta_t$ and $d_t$ are small, $\phi_t$ is small and can be approximated as:
\begin{equation}
\begin{aligned}
\phi_t \approx \tan(\phi) \approx \frac{\| H(\theta_t)d_t \|}{\| \nabla f(\theta_t) \|} \le L \eta_t.
\end{aligned}
\label{phi}
\end{equation}
Plug Eq. (\ref{phi}) into Eq. (\ref{Exp_2_2}), we have:
\begin{equation}
\begin{aligned}
RHS\le LG\rho_t \eta_t + L^3 \rho_t^2 \eta_t^2
\end{aligned}
\label{Exp_2_3}
\end{equation}
Combine Eq. (\ref{Exp_2_3}) and Eq. (\ref{Exp_2_1}), we have:
\begin{equation}
\begin{aligned}
\eta_t\| \nabla f(\theta_t^{sam}) \|^2 &\le f(\theta_t^{sam}) - \mathbb{E}[f(\theta_{t+1}^{sam})] + 2 L G \rho_t \eta_t \\
& + L G^2 \eta_t^2 + L^3 \rho_t^2 \eta_t^2
\end{aligned}
\label{Exp_2_4}
\end{equation}
By setting $\eta_t = \frac{\eta_0}{\sqrt{t}}$ and $\rho_t = \frac{\rho_0}{\sqrt{t}}$, we perform telescope sum and take the expectation on each step $t$. Then we have:
\begin{equation}
\begin{aligned}
\sum_{t=1}^T \eta_t\| \nabla f(\theta_t^{sam}) \|^2 &\le f(\theta_1^{sam}) - \mathbb{E}[f(\theta_T^{sam})] + L^3 \rho_0^2 \eta_0^2 \sum_{t=1}^T \frac{1}{t^2} \\
& + (2LG\rho_0\eta_0 + LG^2\eta_0^2)\sum_{t=1}^T \frac{1}{t} \\
\frac{\eta_0}{\sqrt{T}} \sum_{t=1}^T \| \nabla f(\theta_t^{sam}&) \|^2 \le f(\theta_1^{sam}) - f_{min} + \frac{\pi^2 L^3 \rho_0^2 \eta_0^2}{6} \\
& + (2LG\rho_0\eta_0 + LG^2\eta_0^2)(1+\log T),
\end{aligned}
\label{telescope_2_1}
\end{equation}
Hence
\begin{equation}
\begin{aligned}
\frac{1}{T} \sum_{t=1}^T \| \nabla f(\theta_t^{sam}) \|^2 \le 
\frac{C_3 + C_4 \log T}{\sqrt{T}}
\end{aligned}
\label{Convergence_2}
\end{equation}
implies the convergence rate w.r.t. $f(\theta^{sam})$ is $O(\frac{\log T}{\sqrt{T}})$, where $C_3$ and $C_4$ are some constants.

\subsubsection{Convergence W.R.T. Function $f(\theta^{meta})$}
\label{convergence_meta}
Denote the update $d_t$ of the model parameter $\theta$ at step $t$ as: 
\begin{equation}
\begin{aligned}
d_t = -\eta_t(\gamma g_t+\alpha g_t^{sam}+\alpha g_t^{meta}), \gamma = 1 - 2\alpha.
\end{aligned}
\label{dt_3}
\end{equation}
Define $d_t=\theta_{t+1}-\theta_t$, $f(\theta_t^{sam}) = f(\theta_t + \delta_t)$, and $f(\theta_t^{meta}) = f(\theta_t - \delta_t)$.
By $L$-smoothness of $f$, we have:
\begin{equation}
\begin{aligned}
f(\theta_{t+1}^{meta}) &\le f(\theta_t^{meta}) + \langle \nabla f(\theta_t^{sam}), \theta_{t+1}^{meta}-\theta_t^{meta} \rangle \\
&+\frac{L}{2}\| \theta_{t+1}^{meta}-\theta_t^{meta} \|^2 \\
&=f(\theta_t^{meta}) + \langle \nabla f(\theta_t^{meta}), \theta_{t+1}-\theta_t+\delta_t-\delta_{t+1} \rangle \\
&+\frac{L}{2}\| \theta_{t+1}-\theta_t+\delta_t-\delta_{t+1} \|^2.
\end{aligned}
\label{Lipschitz_3_1}
\end{equation}
We re-arrange the above formula and then obtain:
\begin{equation}
\begin{aligned}
& f(\theta_{t+1}^{meta}) - f(\theta_t^{meta}) - \langle \nabla f(\theta_t^{meta}), d_t \rangle - L \| d_t \|^2 \le \\
& \langle \nabla f(\theta_t^{meta}), \delta_t-\delta_{t+1} \rangle + L \| \delta_t-\delta_{t+1} \|^2.
\end{aligned}
\label{Lipschitz_2_2}
\end{equation}
Similar to the proof in \ref{convergence_sam}, we take the expectation conditioned on the observations up to step $t$ for the both sides and obtain:
\begin{equation}
\begin{aligned}
\mathbb{E}[f(\theta_{t+1}^{meta})] &- f(\theta_t^{meta}) + \eta_t\| \nabla f(\theta_t^{meta}) \|^2 \\
- L G^2 \eta_t^2 &\le L^3 \rho_t^2 \eta_t^2 \\
\eta_t\| \nabla f(\theta_t^{meta}) \|^2 &\le f(\theta_t^{meta}) - \mathbb{E}[f(\theta_{t+1}^{meta})] \\
& + L G^2 \eta_t^2 + L^3 \rho_t^2 \eta_t^2.
\end{aligned}
\label{Exp_3}
\end{equation}
Following Eq. (\ref{telescope_2_1}) and Eq. (\ref{Convergence_2}), we have:
\begin{equation}
\begin{aligned}
\frac{1}{T} \sum_{t=1}^T \| \nabla f(\theta_t^{meta}) \|^2 \le 
\frac{C_5 + C_6 \log T}{\sqrt{T}}, 
\end{aligned}
\label{Convergence_3}
\end{equation}
where $C_5$ and $C_6$ are some constants. Eq. (\ref{Convergence_3}) implies the convergence rate w.r.t. $f(\theta^{meta})$ is $O(\frac{\log T}{\sqrt{T}})$.

\subsubsection{Convergence W.R.T. Overall Function $F(\theta)$}
\label{convergence_overall}
Note that we have proved convergence for $f(\theta)$, $f(\theta^{sam})$, and $f(\theta^{meta})$ in \ref{convergence_oracle}, \ref{convergence_sam}, and \ref{convergence_meta}, respectively. By definition of $F(\theta)$, we have:
\begin{equation}
\begin{aligned}
\| \nabla F(\theta_t) \|^2 &= \| \gamma \nabla f(\theta_t) + \alpha \nabla f(\theta_t^{sam}) + \alpha \nabla f(\theta_t^{meta}) \|^2 \\
&\le \frac{10\gamma^2}{3} \| \nabla f(\theta_t) \|^2 + \frac{10\alpha^2}{3} \| \nabla f(\theta_t^{sam}) \|^2 \\
&+ \frac{10\alpha^2}{3} \| \nabla f(\theta_t^{meta}) \|^2.
\end{aligned}
\label{overall_1}
\end{equation}
Hence
\begin{equation}
\begin{aligned}
&\frac{1}{T} \sum_{t=1}^T \| \nabla F(\theta_t) \|^2 \le \frac{10}{3T}(\gamma^2 \sum_{t=1}^T \| \nabla f(\theta_t) \|^2 + \\
&\alpha^2 \sum_{t=1}^T \| \nabla f(\theta_t^{sam}) \|^2 
 + \alpha^2 \sum_{t=1}^T \| \nabla f(\theta_t^{meta}) \|^2)
\end{aligned}
\label{overall_1}
\end{equation}
also converges at rate $O(\frac{\log T}{\sqrt{T}})$ since each item in the $RHS$ converges at rate $O(\frac{\log T}{\sqrt{T}})$.

\begin{table*}[!htb]
    \caption{The optimal hyperparameter configurations for MeCAM on each dataset. The $\rho$, $\alpha$, and $\beta$ denote the perturbation radius, the weight for the surrogate gap of SAM, and the weight for the surrogate gap of meta-learning, respectively.
    }
    \centering
    \resizebox{0.7\textwidth}{!}{
    \begin{tabular}{c|cccccc}
        \Xhline{1pt}
        Dataset &
        learning rate &
        weight decay &
        dropout rate &
        $\rho$ &
        $\alpha$ &
        $\beta$ \\
        \hline
        PACS &
        $3e^{-5}$ &
        $1e^{-4}$ &
        $0.5$ &
        $0.1$ &
        $0.1$ &
        $0.1$ \\
        VLCS &
        $1e^{-6}$ &
        $1e^{-4}$ &
        $0.5$ &
        $0.1$ &
        $0.2$ &
        $0.1$ \\
        OfficeHome &
        $1e^{-5}$ &
        $1e^{-4}$ &
        $0.5$ &
        $0.2$ &
        $0.2$ &
        $0.1$ \\
        TerraIncognita &
        $1e^{-5}$ &
        $1e^{-4}$ &
        $0.5$ &
        $0.01$ &
        $0.05$ &
        $0.05$ \\
        DomainNet &
        $3e^{-5}$ &
        $1e^{-6}$ &
        $0.5$ &
        $0.1$ &
        $0.2$ &
        $0.1$ \\
        \Xhline{1pt}
    \end{tabular}
    }
    \label{tab:hyperparameter}
\end{table*}

\subsection{Proof of Proposition 3.5} 
By performing first-order Taylor expansion around $\theta$, we have:
\begin{equation}
\begin{aligned}
f(\theta-\delta) \approx f(\theta)-\delta \nabla f(\theta), \\
m(\theta-\delta) \approx m(\theta)-\delta \nabla m(\theta),
\label{eq:meta_Taylor}
\end{aligned}
\end{equation}
where the remainders are omitted. By assuming this proposition holds, we have:
\begin{equation}
\begin{aligned}
\alpha (f(\theta-\delta) - f(\theta)) \approx \beta (m(\theta-\delta) - f(\theta)).
\label{eq:proposition35}
\end{aligned}
\end{equation}
Applying Eq. (\ref{eq:meta_Taylor}) to Eq. (\ref{eq:proposition35}), when the perturbation $\delta$ is small, we can obtain:
\begin{equation}
\begin{aligned}
-\alpha \delta \nabla f(\theta) &\approx \beta (m(\theta) - f(\theta)) - \beta \delta \nabla m(\theta) \\
f(\theta) - m(\theta) &\approx \delta(\frac{\alpha}{\beta} \nabla f(\theta)-\nabla m(\theta))\\
\beta &\approx \alpha \frac{\nabla f(\theta)}{\nabla m(\theta)+\frac{C^\prime}{\delta}},
\end{aligned}
\end{equation}
where $f(\theta) - m(\theta) \approx C^\prime$ and $C^\prime$ is a small constant. 
For convenience, we set $\beta$ as a hyperparameter, which meets the $\beta \le \alpha$ condition.

\section{Optimal Hyperparameter Configurations}
\label{Hyperparameter}
For a fair comparison, we followed the training and evaluation protocol in~\cite{gulrajanisearch} to search for the optimal hyperparameters, where the test data information is unavailable. The search spaces of the learning rate (\emph{i.e.}, `lr'), the perturbation radius $\rho$, the weight for the surrogate gap of SAM $\alpha$, and the weight for the surrogate gap of meta-learning $\beta$ are $\{1e^{-6}, 1e^{-5}, 3e^{-5}\}$, $\{0.01, 0.05, 0.1, 0.2\}$, $\{0.05, 0.1, 0.15, 0.2\}$, and $\{0.05, 0.1, 0.15, 0.2\}$, respectively.
For the weight decay (\emph{i.e.}, `wd') and the dropout rate (\emph{i.e.}, `dr'), we inherited the optimal configurations in~\cite{SAGM}.
The optimal hyperparameter configurations for MeCAM on each dataset are provided in Table~\ref{tab:hyperparameter}.
It is worth noting that all the optimal configurations also satisfy the condition $\beta \le \alpha$.

\section{Full Results}
\label{Full_Result}
In this section, we provide the detailed results of Table 2 and Table 3 shown in our manuscript.
Specifically, we display the full results of our MeCAM and other existing DG methods~\cite{MMD,MixStyle,GroupDRO,IRM,ARM,VREx,AND-mask,CDANN,DANN,MTL,Mixup,MLDG,ERM,SagNet,Fishr,MADG} on PACS, VLCS, OfficeHome, TerraIncognita, and DomainNet datasets in Table~\ref{tab:PACS}, Table~\ref{tab:VLCS}, Table~\ref{tab:OfficeHome}, Table~\ref{tab:TerraIncognita}, and Table~\ref{tab:DomainNet}, respectively.
We also display the full results of our MeCAM, optimizers~\cite{Adam,AdamW,SGD,YOGI,AdaBelief,AdaHessian}, and existing sharpness-based DG methods~\cite{SAM,GAM,SAGM,FAD,CRSAM,FSAM} on PACS, VLCS, OfficeHome, TerraIncognita, and DomainNet datasets in Table~\ref{tab:PACS2}, Table~\ref{tab:VLCS2}, Table~\ref{tab:OfficeHome2}, Table~\ref{tab:TerraIncognita2}, and Table~\ref{tab:DomainNet2}, respectively.
For the results of each method, we report the average accuracy and standard deviation computed over three trials using random seeds of 0, 1, and 2.


\begin{table*}[!h]
    \caption{Full results ($mean_{\pm std}$) of our MeCAM and existing DG methods calculated across three trials on PACS. The results denoted by $\dag$ are taken from~\cite{SAGM} and~\cite{MADG}, while the results marked by $\ddag$ are inherited directly from the original source.}
    \centering
    \resizebox{0.95\textwidth}{!}{
    \begin{tabular}{c|cccc|c}
        \Xhline{1pt}
        Algorithm &
        Art &
        Cartoon &
        Photo &
        Sketch &
        Average \\
        \hline
        MMD$^\dag$~\cite{MMD} & 
        $86.1_{\pm1.4}$ &
        $79.4_{\pm0.9}$ &
        $96.6_{\pm0.2}$ &
        $76.5_{\pm0.5}$ &
        $84.7$ \\
        Mixstyle$^\dag$~\cite{MixStyle} & 
        $86.8_{\pm0.5}$ &
        $79.0_{\pm1.4}$ &
        $96.6_{\pm0.1}$ &
        $78.5_{\pm2.3}$ &
        $85.2$ \\
        GroupDRO$^\dag$~\cite{GroupDRO} & 
        $83.5_{\pm0.9}$ &
        $79.1_{\pm0.6}$ &
        $96.7_{\pm0.3}$ &
        $78.3_{\pm2.0}$ &
        $84.4$ \\
        IRM$^\dag$~\cite{IRM} & 
        $84.8_{\pm1.3}$ &
        $76.4_{\pm1.1}$ &
        $96.7_{\pm0.6}$ &
        $76.1_{\pm1.0}$ &
        $83.5$ \\
        ARM$^\dag$~\cite{ARM} & 
        $86.8_{\pm0.6}$ &
        $76.8_{\pm0.5}$ &
        $97.4_{\pm0.3}$ &
        $79.3_{\pm1.2}$ &
        $85.1$ \\
        VREx$^\dag$~\cite{VREx} & 
        $86.0_{\pm1.6}$ &
        $79.1_{\pm0.6}$ &
        $96.9_{\pm0.5}$ &
        $77.7_{\pm1.7}$ &
        $84.9$ \\
        AND-mask$^\ddag$~\cite{SAND-mask} &
        $86.4_{\pm1.1}$ &
        $80.8_{\pm0.9}$ &
        $97.1_{\pm0.2}$ &
        $81.3_{\pm1.1}$ &
        $86.4$ \\
        CDANN$^\dag$~\cite{CDANN} & 
        $84.6_{\pm1.8}$ &
        $75.5_{\pm0.9}$ &
        $96.8_{\pm0.3}$ &
        $73.5_{\pm0.6}$ &
        $82.6$ \\
        SAND-mask$^\ddag$~\cite{SAND-mask} & 
        $86.1_{\pm0.6}$ &
        $80.3_{\pm1.0}$ &
        $97.1_{\pm0.3}$ &
        $80.0_{\pm1.3}$ &
        $85.9$ \\
        DANN$^\dag$~\cite{DANN} & 
        $86.4_{\pm0.8}$ &
        $77.4_{\pm0.8}$ &
        $97.3_{\pm0.4}$ &
        $73.5_{\pm2.3}$ &
        $83.7$ \\
        MTL$^\dag$~\cite{MTL} & 
        $87.5_{\pm0.8}$ &
        $77.1_{\pm0.5}$ &
        $96.4_{\pm0.8}$ &
        $77.3_{\pm1.8}$ &
        $84.6$ \\
        Mixup$^\dag$~\cite{Mixup} & 
        $86.1_{\pm0.5}$ &
        $78.9_{\pm0.8}$ &
        $97.6_{\pm0.1}$ &
        $75.8_{\pm1.8}$ &
        $84.6$ \\
        MLDG$^\dag$~\cite{MLDG} & 
        $85.5_{\pm1.4}$ &
        $80.1_{\pm1.7}$ &
        $97.4_{\pm0.3}$ &
        $76.6_{\pm1.1}$ &
        $84.9$ \\
        ERM$^\dag$~\cite{ERM} & 
        $84.7_{\pm0.4}$ &
        $80.8_{\pm0.6}$ &
        $97.2_{\pm0.3}$ &
        $79.3_{\pm1.0}$ &
        $85.5$ \\
        SagNet$^\dag$~\cite{SagNet} & 
        $87.4_{\pm1.0}$ &
        $80.7_{\pm0.6}$ &
        $97.1_{\pm0.1}$ &
        $80.0_{\pm0.4}$ &
        $86.3$ \\
        Fishr$^\dag$~\cite{Fishr} &
        $87.9_{\pm0.6}$ &
        $80.8_{\pm0.5}$ &
        $97.9_{\pm0.4}$ &
        $81.1_{\pm0.8}$ &
        $86.9$ \\
        MADG$^\dag$~\cite{MADG} &
        $87.8_{\pm0.5}$ &
        $82.2_{\pm0.6}$ &
        $97.7_{\pm0.3}$ &
        $78.3_{\pm0.4}$ &
        $86.5$ \\
        \hline
        MeCAM (Ours) & 
        $89.6_{\pm0.6}$ &
        $83.3_{\pm1.4}$ &
        $96.7_{\pm0.3}$ &
        $83.5_{\pm0.7}$ &
        $88.3$ \\ 
        \Xhline{1pt}
    \end{tabular}
    }
    \label{tab:PACS}
\end{table*}

\begin{table*}[!h]
    \caption{Full results ($mean_{\pm std}$) of our MeCAM, optimizers, and existing sharpness-based DG methods calculated across three trials on PACS. Results marked with $\dag$ and * are inherited from~\cite{FAD} and obtained by reproduction, respectively. }
    \centering
    \resizebox{\textwidth}{!}{
    \begin{tabular}{c|c|cccc|c}
        \Xhline{1pt}
        \multicolumn{2}{c|}{Algorithm} & 
        Art &
        Cartoon &
        Photo &
        Sketch &
        Average \\
        \hline
        {\multirow{6}*{Optimizer}}
        & Adam$^\dag$~\cite{Adam} &
        $88.0_{\pm1.2}$ &
        $79.7_{\pm0.5}$ &
        $96.7_{\pm0.4}$ &
        $72.7_{\pm0.9}$ &
        $84.3$ \\
        & AdamW$^\dag$~\cite{AdamW} &
        $84.1_{\pm1.5}$ &
        $80.7_{\pm1.2}$ &
        $96.9_{\pm0.4}$ &
        $72.8_{\pm0.6}$ &
        $83.6$ \\
        & SGD$^\dag$~\cite{SGD}	&
        $85.1_{\pm0.4}$ &
        $76.0_{\pm0.3}$ &
        $98.3_{\pm0.4}$ &
        $60.3_{\pm6.1}$ &
        $79.9$ \\
        & YOGI$^\dag$~\cite{YOGI} &
        $84.4_{\pm1.7}$ &
        $79.7_{\pm0.6}$ &
        $95.8_{\pm0.3}$ &
        $65.1_{\pm1.5}$ &
        $81.2$ \\
        & AdaBelief$^\dag$~\cite{AdaBelief}	&
        $85.4_{\pm2.2}$ &
        $80.4_{\pm1.1}$ &
        $97.4_{\pm0.7}$ &
        $75.1_{\pm1.4}$ &
        $84.6$ \\
        & AdaHessian$^\dag$~\cite{AdaHessian}	&
        $88.4_{\pm0.6}$ &
        $80.0_{\pm0.9}$ &
        $97.7_{\pm0.4}$ &
        $71.7_{\pm4.1}$ &
        $84.5$ \\
        \hline
        {\multirow{6}*{Sharpness-based}}
        & SAM$^\dag$~\cite{SAM}	&
        $85.7_{\pm1.2}$ &
        $81.0_{\pm1.4}$ &
        $97.1_{\pm0.2}$ &
        $77.4_{\pm1.8}$ &
        $85.3$ \\
        & GAM$^\dag$~\cite{GAM}	&
        $85.9_{\pm0.9}$ &
        $81.3_{\pm1.6}$ &
        $98.2_{\pm0.4}$ &
        $79.0_{\pm2.1}$ &
        $86.1$ \\
        & SAGM*~\cite{SAGM}	&
        $87.6_{\pm0.4}$ &
        $81.8_{\pm0.5}$ &
        $97.2_{\pm0.1}$ &
        $81.3_{\pm2.1}$ &
        $86.9$ \\
        & FAD$^\dag$~\cite{FAD}	&
        $88.5_{\pm0.5}$ &
        $83.0_{\pm0.8}$ &
        $98.4_{\pm0.2}$ &
        $82.8_{\pm0.9}$ &
        $88.2$ \\
        & CRSAM*~\cite{CRSAM} &
        $86.7_{\pm0.5}$ &
        $80.5_{\pm1.6}$ &
        $96.4_{\pm0.5}$ &
        $77.8_{\pm2.4}$ &
        $85.4$ \\
        & FSAM*~\cite{FSAM}	&
        $87.6_{\pm1.1}$ &
        $80.2_{\pm0.7}$ &
        $95.4_{\pm0.4}$ &
        $82.8_{\pm0.7}$ &
        $86.5$ \\
        \hline
        \multicolumn{2}{c|}{MeCAM (Ours)} & 
        $89.6_{\pm0.6}$ &
        $83.3_{\pm1.4}$ &
        $96.7_{\pm0.3}$ &
        $83.5_{\pm0.7}$ &
        $88.3$ \\ 
        \Xhline{1pt}
    \end{tabular}
    }
    \label{tab:PACS2}
\end{table*}


\begin{table*}[!h]
    \caption{Full results ($mean_{\pm std}$) of our MeCAM and existing DG methods calculated across three trials on VLCS. The results denoted by $\dag$ are taken from~\cite{SAGM} and~\cite{MADG}, while the results marked by $\ddag$ are inherited directly from the original source.}
    \centering
    \resizebox{0.95\textwidth}{!}{
    \begin{tabular}{c|cccc|c}
        \Xhline{1pt}
        Algorithm &
        Caltech &
        LabelMe &
        SUN &
        VOC &
        Average \\
        \hline
        MMD$^\dag$~\cite{MMD} & 
        $97.7_{\pm0.1}$ &
        $64.0_{\pm1.1}$ &
        $72.8_{\pm0.2}$ &
        $75.3_{\pm3.3}$ &
        $77.5$ \\ 
        Mixstyle$^\dag$~\cite{MixStyle} & 
        $98.6_{\pm0.3}$ &
        $64.5_{\pm1.1}$ &
        $72.6_{\pm0.5}$ &
        $75.7_{\pm1.7}$ &
        $77.9$ \\ 
        GroupDRO$^\dag$~\cite{GroupDRO} & 
        $97.3_{\pm0.3}$ &
        $63.4_{\pm0.9}$ &
        $69.5_{\pm0.8}$ &
        $76.7_{\pm0.7}$ &
        $76.7$ \\ 
        IRM$^\dag$~\cite{IRM} & 
        $98.6_{\pm0.1}$ &
        $64.9_{\pm0.9}$ &
        $73.4_{\pm0.6}$ &
        $77.3_{\pm0.9}$ &
        $78.6$ \\ 
        ARM$^\dag$~\cite{ARM} & 
        $98.7_{\pm0.2}$ &
        $63.6_{\pm0.7}$ &
        $71.3_{\pm1.2}$ &
        $76.7_{\pm0.6}$ &
        $77.6$ \\ 
        VREx$^\dag$~\cite{VREx} & 
        $98.4_{\pm0.3}$ &
        $64.4_{\pm1.4}$ &
        $74.1_{\pm0.4}$ &
        $76.2_{\pm1.3}$ &
        $78.3$ \\ 
        AND-mask$^\ddag$~\cite{SAND-mask} &
        $98.3_{\pm0.3}$ &
        $64.5_{\pm0.2}$ &
        $69.3_{\pm1.3}$ &
        $73.4_{\pm1.3}$ &
        $76.4$ \\ 
        CDANN$^\dag$~\cite{CDANN} & 
        $97.1_{\pm0.3}$ &
        $65.1_{\pm1.2}$ &
        $70.7_{\pm0.8}$ &
        $77.1_{\pm1.5}$ &
        $77.5$ \\ 
        SAND-mask$^\ddag$~\cite{SAND-mask} & 
        $97.6_{\pm0.3}$ &
        $64.5_{\pm0.6}$ &
        $69.7_{\pm0.6}$ &
        $73.0_{\pm1.2}$ &
        $76.2$ \\
        DANN$^\dag$~\cite{DANN} & 
        $99.0_{\pm0.3}$ &
        $65.1_{\pm1.4}$ &
        $73.1_{\pm0.3}$ &
        $77.2_{\pm0.6}$ &
        $78.6$ \\
        MTL$^\dag$~\cite{MTL} & 
        $97.8_{\pm0.4}$ &
        $64.3_{\pm0.3}$ &
        $71.5_{\pm0.7}$ &
        $75.3_{\pm1.7}$ &
        $77.2$ \\
        Mixup$^\dag$~\cite{Mixup} & 
        $98.3_{\pm0.6}$ &
        $64.8_{\pm1.0}$ &
        $72.1_{\pm0.5}$ &
        $74.3_{\pm0.8}$ &
        $77.4$ \\
        MLDG$^\dag$~\cite{MLDG} & 
        $97.4_{\pm0.2}$ &
        $65.2_{\pm0.7}$ &
        $71.0_{\pm1.4}$ &
        $75.3_{\pm1.0}$ &
        $77.2$ \\
        ERM$^\dag$~\cite{ERM} & 
        $98.0_{\pm0.3}$ &
        $64.7_{\pm1.2}$ &
        $71.4_{\pm1.2}$ &
        $75.2_{\pm1.6}$ &
        $77.3$ \\
        SagNet$^\dag$~\cite{SagNet} & 
        $97.9_{\pm0.4}$ &
        $64.5_{\pm0.5}$ &
        $71.4_{\pm1.3}$ &
        $77.5_{\pm0.5}$ &
        $77.8$ \\
        Fishr$^\dag$~\cite{Fishr} &
        $97.6_{\pm0.7}$ &
        $67.3_{\pm0.5}$ &
        $72.2_{\pm0.9}$ &
        $75.7_{\pm0.3}$ &
        $78.2$ \\
        MADG$^\dag$~\cite{MADG} &
        $98.5_{\pm0.2}$ &
        $65.8_{\pm0.3}$ &
        $73.1_{\pm0.3}$ &
        $77.3_{\pm0.1}$ &
        $78.7$ \\
        \hline
        MeCAM (Ours) & 
        $99.5_{\pm0.0}$ &
        $64.3_{\pm0.5}$ &
        $75.2_{\pm0.8}$ &
        $82.1_{\pm0.3}$ &
        $80.3$ \\
        \Xhline{1pt}
    \end{tabular}
    }
    \label{tab:VLCS}
\end{table*}

\begin{table*}[!h]
    \caption{Full results ($mean_{\pm std}$) of our MeCAM, optimizers, and existing sharpness-based DG methods calculated across three trials on VLCS. Results marked with $\dag$ and * are inherited from~\cite{FAD} and obtained by reproduction, respectively.
    }
    \centering
    \resizebox{\textwidth}{!}{
    \begin{tabular}{c|c|cccc|c}
        \Xhline{1pt}
        \multicolumn{2}{c|}{Algorithm} & 
        Caltech &
        LabelMe &
        SUN &
        VOC &
        Average \\
        \hline
        {\multirow{6}*{Optimizer}}
        & Adam$^\dag$~\cite{Adam} &
        $98.9_{\pm0.4}$ &
        $65.9_{\pm1.5}$ &
        $71.0_{\pm1.6}$ &
        $74.5_{\pm2.0}$ &
        $77.3$ \\
        & AdamW$^\dag$~\cite{AdamW} &
        $98.3_{\pm0.1}$ &
        $65.1_{\pm1.7}$ &
        $70.9_{\pm1.3}$ &
        $75.2_{\pm1.5}$ &
        $77.4$ \\
        & SGD$^\dag$~\cite{SGD}	&
        $98.4_{\pm0.2}$ &
        $64.7_{\pm0.7}$ &
        $72.5_{\pm0.8}$ &
        $76.6_{\pm0.8}$ &
        $78.1$ \\
        & YOGI$^\dag$~\cite{YOGI} &
        $98.1_{\pm0.7}$ &
        $63.9_{\pm1.2}$ &
        $72.5_{\pm1.6}$ &
        $75.7_{\pm1.2}$ &
        $77.6$ \\
        & AdaBelief$^\dag$~\cite{AdaBelief}	&
        $98.0_{\pm0.1}$ &
        $63.9_{\pm0.4}$ &
        $73.4_{\pm1.0}$ &
        $78.2_{\pm1.8}$ &
        $78.4$ \\
        & AdaHessian$^\dag$~\cite{AdaHessian} &
        $99.1_{\pm0.3}$ &
        $65.0_{\pm1.7}$ &
        $72.7_{\pm1.3}$ &
        $77.7_{\pm1.0}$ &
        $78.6$ \\
        \hline
        {\multirow{6}*{Sharpness-based}}
        & SAM$^\dag$~\cite{SAM}	&
        $98.5_{\pm1.0}$ &
        $66.2_{\pm1.6}$ &
        $72.0_{\pm1.0}$ &
        $76.1_{\pm1.0}$ &
        $78.2$ \\
        & GAM$^\dag$~\cite{GAM}	&
        $98.8_{\pm0.6}$ &
        $65.1_{\pm1.2}$ &
        $72.9_{\pm1.0}$ &
        $77.2_{\pm1.9}$ &
        $78.5$ \\
        & SAGM*~\cite{SAGM}	&
        $98.4_{\pm0.6}$ &
        $65.1_{\pm1.1}$ &
        $74.1_{\pm0.5}$ &
        $78.6_{\pm2.4}$ &
        $79.1$ \\ 
        & FAD$^\dag$~\cite{FAD}	&
        $99.1_{\pm0.5}$ &
        $66.8_{\pm0.9}$ &
        $73.6_{\pm1.0}$ &
        $76.1_{\pm1.3}$ &
        $78.9$ \\
        & CRSAM*~\cite{CRSAM} &
        $99.0_{\pm0.4}$ &
        $63.7_{\pm0.8}$ &
        $73.8_{\pm0.5}$ &
        $79.8_{\pm1.4}$ &
        $79.1$ \\ 
        & FSAM*~\cite{FSAM}	&
        $98.9_{\pm0.2}$ &
        $64.5_{\pm0.4}$ &
        $74.1_{\pm1.0}$ &
        $80.0_{\pm1.0}$ &
        $79.4$ \\ 
        \hline
        \multicolumn{2}{c|}{MeCAM (Ours)} & 
        $99.5_{\pm0.0}$ &
        $64.3_{\pm0.5}$ &
        $75.2_{\pm0.8}$ &
        $82.1_{\pm0.3}$ &
        $80.3$ \\
        \Xhline{1pt}
    \end{tabular}
    }
    \label{tab:VLCS2}
\end{table*}

\begin{table*}[!h]
    \caption{Full results ($mean_{\pm std}$) of our MeCAM and existing DG methods calculated across three trials on OfficeHome. The results denoted by $\dag$ are taken from~\cite{SAGM} and~\cite{MADG}, while the results marked by $\ddag$ are inherited directly from the original source.}
    \centering
    \resizebox{0.95\textwidth}{!}{
    \begin{tabular}{c|cccc|c}
        \Xhline{1pt}
        Algorithm &
        Art &
        Clipart &
        Product &
        Real-World &
        Average \\
        \hline
        MMD$^\dag$~\cite{MMD} & 
        $60.4_{\pm0.2}$ &
        $53.3_{\pm0.3}$ &
        $74.3_{\pm0.1}$ &
        $77.4_{\pm0.6}$ &
        $66.4$ \\
        Mixstyle$^\dag$~\cite{MixStyle} & 
        $51.1_{\pm0.3}$ &
        $53.2_{\pm0.4}$ &
        $68.2_{\pm0.7}$ &
        $69.2_{\pm0.6}$ &
        $60.4$ \\
        GroupDRO$^\dag$~\cite{GroupDRO} & 
        $60.4_{\pm0.7}$ &
        $52.7_{\pm1.0}$ &
        $75.0_{\pm0.7}$ &
        $76.0_{\pm0.7}$ &
        $66.0$ \\
        IRM$^\dag$~\cite{IRM} & 
        $58.9_{\pm2.3}$ &
        $52.2_{\pm1.6}$ &
        $72.1_{\pm2.9}$ &
        $74.0_{\pm2.5}$ &
        $64.3$ \\
        ARM$^\dag$~\cite{ARM} & 
        $58.9_{\pm0.8}$ &
        $51.0_{\pm0.5}$ &
        $74.1_{\pm0.1}$ &
        $75.2_{\pm0.3}$ &
        $64.8$ \\
        VREx$^\dag$~\cite{VREx} & 
        $60.7_{\pm0.9}$ &
        $53.0_{\pm0.9}$ &
        $75.3_{\pm0.1}$ &
        $76.6_{\pm0.5}$ &
        $66.4$ \\
        AND-mask$^\ddag$~\cite{SAND-mask} &
        $60.3_{\pm0.5}$ &
        $52.3_{\pm0.6}$ &
        $75.1_{\pm0.2}$ &
        $76.6_{\pm0.3}$ &
        $66.1$ \\
        CDANN$^\dag$~\cite{CDANN} & 
        $61.5_{\pm1.4}$ &
        $50.4_{\pm2.4}$ &
        $74.4_{\pm0.9}$ &
        $76.6_{\pm0.8}$ &
        $65.7$ \\
        SAND-mask$^\ddag$~\cite{SAND-mask} & 
        $59.9_{\pm0.7}$ &
        $53.6_{\pm0.8}$ &
        $74.3_{\pm0.4}$ &
        $75.8_{\pm0.5}$ &
        $65.9$ \\
        DANN$^\dag$~\cite{DANN} & 
        $59.9_{\pm1.3}$ &
        $53.0_{\pm0.3}$ &
        $73.6_{\pm0.7}$ &
        $76.9_{\pm0.5}$ &
        $65.9$ \\
        MTL$^\dag$~\cite{MTL} & 
        $61.5_{\pm0.7}$ &
        $52.4_{\pm0.6}$ &
        $74.9_{\pm0.4}$ &
        $76.8_{\pm0.4}$ &
        $66.4$ \\
        Mixup$^\dag$~\cite{Mixup} & 
        $62.4_{\pm0.8}$ &
        $54.8_{\pm0.6}$ &
        $76.9_{\pm0.3}$ &
        $78.3_{\pm0.2}$ &
        $68.1$ \\
        MLDG$^\dag$~\cite{MLDG} & 
        $61.5_{\pm0.9}$ &
        $53.2_{\pm0.6}$ &
        $75.0_{\pm1.2}$ &
        $77.5_{\pm0.4}$ &
        $66.8$ \\
        ERM$^\dag$~\cite{ERM} & 
        $61.3_{\pm0.7}$ &
        $52.4_{\pm0.3}$ &
        $75.8_{\pm0.1}$ &
        $76.6_{\pm0.3}$ &
        $66.5$ \\
        SagNet$^\dag$~\cite{SagNet} & 
        $63.4_{\pm0.2}$ &
        $54.8_{\pm0.4}$ &
        $75.8_{\pm0.4}$ &
        $78.3_{\pm0.3}$ &
        $68.1$ \\
        Fishr$^\dag$~\cite{Fishr} &
        $63.4_{\pm0.8}$ &
        $54.2_{\pm0.3}$ &
        $76.4_{\pm0.3}$ &
        $78.5_{\pm0.2}$ &
        $68.2$ \\
        MADG$^\dag$~\cite{MADG} &
        $68.6_{\pm0.5}$ &
        $55.5_{\pm0.2}$ &
        $79.6_{\pm0.3}$ &
        $81.5_{\pm0.4}$ &
        $71.3$ \\
        \hline
        MeCAM (Ours) & 
        $65.9_{\pm0.6}$ &
        $58.7_{\pm0.5}$ &
        $77.1_{\pm0.5}$ &
        $79.9_{\pm0.3}$ &
        $70.4$ \\ 
        \Xhline{1pt}
    \end{tabular}
    }
    \label{tab:OfficeHome}
\end{table*}

\begin{table*}[!h]
    \caption{Full results ($mean_{\pm std}$) of our MeCAM, optimizers, and existing sharpness-based DG methods calculated across three trials on OfficeHome. Results marked with $\dag$ and * are inherited from~\cite{FAD} and obtained by reproduction, respectively.
    }
    \centering
    \resizebox{\textwidth}{!}{
    \begin{tabular}{c|c|cccc|c}
        \Xhline{1pt}
        \multicolumn{2}{c|}{Algorithm} & 
        Art &
        Clipart &
        Product &
        Real-World &
        Average \\
        \hline
        {\multirow{6}*{Optimizer}}
        & Adam$^\dag$~\cite{Adam} &
        $63.9_{\pm0.8}$ &
        $48.1_{\pm0.6}$ &
        $77.0_{\pm0.9}$ &
        $81.8_{\pm1.6}$ &
        $67.6$ \\
        & AdamW$^\dag$~\cite{AdamW} &
        $66.1_{\pm0.7}$ &
        $48.7_{\pm0.6}$ &
        $76.6_{\pm0.8}$ &
        $83.6_{\pm0.4}$ &
        $68.8$ \\
        & SGD$^\dag$~\cite{SGD}	&
        $65.3_{\pm0.8}$ &
        $48.8_{\pm1.4}$ &
        $76.7_{\pm0.3}$ &
        $83.0_{\pm0.7}$ &
        $68.5$ \\
        & YOGI$^\dag$~\cite{YOGI} &
        $63.5_{\pm1.0}$ &
        $49.2_{\pm1.2}$ &
        $76.2_{\pm0.5}$ &
        $84.5_{\pm0.6}$ &
        $68.3$ \\
        & AdaBelief$^\dag$~\cite{AdaBelief}	&
        $65.6_{\pm2.0}$ &
        $48.1_{\pm0.9}$ &
        $74.8_{\pm0.8}$ &
        $83.6_{\pm0.9}$ &
        $68.0$ \\
        & AdaHessian$^\dag$~\cite{AdaHessian} &
        $63.0_{\pm2.9}$ &
        $50.0_{\pm1.4}$ &
        $77.7_{\pm0.8}$ &
        $83.0_{\pm0.5}$ &
        $68.4$ \\
        \hline
        {\multirow{6}*{Sharpness-based}}
        & SAM$^\dag$~\cite{SAM}	&
        $63.5_{\pm1.2}$ &
        $48.6_{\pm0.9}$ &
        $77.0_{\pm0.8}$ &
        $82.9_{\pm1.3}$ &
        $68.0$ \\
        & GAM$^\dag$~\cite{GAM}	&
        $63.0_{\pm1.2}$ &
        $49.8_{\pm0.5}$ &
        $77.6_{\pm0.6}$ &
        $82.4_{\pm1.0}$ &
        $68.2$ \\
        & SAGM*~\cite{SAGM}	&
        $64.2_{\pm0.6}$ &
        $56.2_{\pm0.4}$ &
        $78.2_{\pm0.2}$ &
        $79.2_{\pm0.1}$ &
        $69.4$ \\ 
        & FAD$^\dag$~\cite{FAD}	&
        $63.5_{\pm1.0}$ &
        $50.3_{\pm0.8}$ &
        $78.0_{\pm0.4}$ &
        $85.0_{\pm0.6}$ &
        $69.2$ \\
        & CRSAM*~\cite{CRSAM} &
        $60.8_{\pm3.7}$ &
        $56.9_{\pm0.4}$ &
        $78.3_{\pm0.5}$ &
        $79.5_{\pm0.7}$ &
        $68.9$ \\ 
        & FSAM*~\cite{FSAM}	&
        $64.7_{\pm0.3}$ &
        $57.3_{\pm0.8}$ &
        $78.8_{\pm0.1}$ &
        $79.9_{\pm0.2}$ &
        $70.2$ \\ 
        \hline
        \multicolumn{2}{c|}{MeCAM (Ours)} & 
        $65.9_{\pm0.6}$ &
        $58.7_{\pm0.5}$ &
        $77.1_{\pm0.5}$ &
        $79.9_{\pm0.3}$ &
        $70.4$ \\ 
        \Xhline{1pt}
    \end{tabular}
    }
    \label{tab:OfficeHome2}
\end{table*}


\begin{table*}[!h]
    \caption{Full results ($mean_{\pm std}$) of our MeCAM and existing DG methods calculated across three trials on TerraIncognita. The results denoted by $\dag$ are taken from~\cite{SAGM} and~\cite{MADG}, while the results marked by $\ddag$ are inherited directly from the original source. 
    }
    \centering
    \resizebox{0.95\textwidth}{!}{
    \begin{tabular}{c|cccc|c}
        \Xhline{1pt}
        Algorithm &
        L100 &
        L38 &
        L43 &
        L46 &        
        Average \\
        \hline
        MMD$^\dag$~\cite{MMD} & 
        $41.9_{\pm3.0}$ &
        $34.8_{\pm1.0}$ &
        $57.0_{\pm1.9}$ &
        $35.2_{\pm1.8}$ &
        $42.2$ \\
        Mixstyle$^\dag$~\cite{MixStyle} & 
        $54.3_{\pm1.1}$ &
        $34.1_{\pm1.1}$ &
        $55.9_{\pm1.1}$ &
        $31.7_{\pm2.1}$ &
        $44.0$ \\
        GroupDRO$^\dag$~\cite{GroupDRO} & 
        $41.2_{\pm0.7}$ &
        $38.6_{\pm2.1}$ &
        $56.7_{\pm0.9}$ &
        $36.4_{\pm2.1}$ &
        $43.2$ \\
        IRM$^\dag$~\cite{IRM} & 
        $54.6_{\pm1.3}$ &
        $39.8_{\pm1.9}$ &
        $56.2_{\pm1.8}$ &
        $39.6_{\pm0.8}$ &
        $47.6$ \\
        ARM$^\dag$~\cite{ARM} & 
        $49.3_{\pm0.7}$ &
        $38.3_{\pm2.4}$ &
        $55.8_{\pm0.8}$ &
        $38.7_{\pm1.3}$ &
        $45.5$ \\
        VREx$^\dag$~\cite{VREx} & 
        $48.2_{\pm4.3}$ &
        $41.7_{\pm1.3}$ &
        $56.8_{\pm0.8}$ &
        $38.7_{\pm3.1}$ &
        $46.4$ \\
        AND-mask$^\ddag$~\cite{SAND-mask} &
        $54.7_{\pm1.8}$ &
        $48.4_{\pm0.5}$ &
        $55.1_{\pm0.5}$ &
        $41.3_{\pm0.6}$ &
        $49.8$ \\
        CDANN$^\dag$~\cite{CDANN} & 
        $47.0_{\pm1.9}$ &
        $41.3_{\pm4.8}$ &
        $54.9_{\pm1.7}$ &
        $39.8_{\pm2.3}$ &
        $45.8$ \\
        SAND-mask$^\ddag$~\cite{SAND-mask} & 
        $56.2_{\pm1.8}$ &
        $46.3_{\pm0.3}$ &
        $55.8_{\pm0.4}$ &
        $42.6_{\pm1.2}$ &
        $50.2$ \\
        DANN$^\dag$~\cite{DANN} & 
        $51.1_{\pm3.5}$ &
        $40.6_{\pm0.6}$ &
        $57.4_{\pm0.5}$ &
        $37.7_{\pm1.8}$ &
        $46.7$ \\
        MTL$^\dag$~\cite{MTL} & 
        $49.3_{\pm1.2}$ &
        $39.6_{\pm6.3}$ &
        $55.6_{\pm1.1}$ &
        $37.8_{\pm0.8}$ &
        $45.6$ \\
        Mixup$^\dag$~\cite{Mixup} & 
        $59.6_{\pm2.0}$ &
        $42.2_{\pm1.4}$ &
        $55.9_{\pm0.8}$ &
        $33.9_{\pm1.4}$ &
        $47.9$ \\ 
        MLDG$^\dag$~\cite{MLDG} & 
        $54.2_{\pm3.0}$ &
        $44.3_{\pm1.1}$ &
        $55.6_{\pm0.3}$ &
        $36.9_{\pm2.2}$ &
        $47.8$ \\
        ERM$^\dag$~\cite{ERM} & 
        $54.3_{\pm0.4}$ &
        $42.5_{\pm0.7}$ &
        $55.6_{\pm0.3}$ &
        $38.8_{\pm2.5}$ &
        $47.8$ \\
        SagNet$^\dag$~\cite{SagNet} & 
        $53.0_{\pm2.9}$ &
        $43.0_{\pm2.5}$ &
        $57.9_{\pm0.6}$ &
        $40.4_{\pm1.3}$ &
        $48.6$ \\
        Fishr$^\dag$~\cite{Fishr} &
        $60.4_{\pm0.9}$ &
        $50.3_{\pm0.3}$ &
        $58.8_{\pm0.5}$ &
        $44.9_{\pm0.5}$ &
        $53.6$ \\
        MADG$^\dag$~\cite{MADG} &
        $60.0_{\pm1.2}$ &
        $51.8_{\pm0.2}$ &
        $57.4_{\pm0.3}$ &
        $45.6_{\pm0.5}$ &
        $53.7$ \\
        \hline
        MeCAM (Ours) & 
        $55.6_{\pm1.4}$ &
        $40.8_{\pm3.6}$ &
        $58.5_{\pm0.5}$ &
        $40.5_{\pm1.6}$ &
        $48.9$ \\
        \Xhline{1pt}
    \end{tabular}
    }
    \label{tab:TerraIncognita}
\end{table*}

\begin{table*}[!h]
    \caption{Full results ($mean_{\pm std}$) of our MeCAM, optimizers, and existing sharpness-based DG methods calculated across three trials on TerraIncognita. Results marked with $\dag$ and * are inherited from~\cite{FAD} and obtained by reproduction, respectively. 
    }
    \centering
    \resizebox{\textwidth}{!}{
    \begin{tabular}{c|c|cccc|c}
        \Xhline{1pt}
        \multicolumn{2}{c|}{Algorithm} & 
        L100 &
        L38 &
        L43 &
        L46 &    
        Average \\
        \hline
        {\multirow{6}*{Optimizer}}
        & Adam$^\dag$~\cite{Adam} &
        $42.2_{\pm3.4}$ &
        $40.7_{\pm1.2}$ &
        $59.9_{\pm0.2}$ &
        $35.0_{\pm2.8}$ &
        $44.4$ \\
        & AdamW$^\dag$~\cite{AdamW} &
        $44.2_{\pm6.8}$ &
        $39.8_{\pm1.9}$ &
        $60.3_{\pm2.0}$ &
        $36.6_{\pm1.8}$ &
        $45.2$ \\
        & SGD$^\dag$~\cite{SGD}	&
        $41.8_{\pm5.8}$ &
        $39.8_{\pm3.9}$ &
        $60.5_{\pm2.2}$ &
        $37.5_{\pm1.1}$ &
        $44.9$ \\
        & YOGI$^\dag$~\cite{YOGI} &
        $43.9_{\pm2.2}$ &
        $42.5_{\pm2.6}$ &
        $60.5_{\pm1.1}$ &
        $34.8_{\pm1.6}$ &
        $45.4$ \\
        & AdaBelief$^\dag$~\cite{AdaBelief}	&
        $42.6_{\pm6.7}$ &
        $43.0_{\pm2.0}$ &
        $60.2_{\pm1.3}$ &
        $35.1_{\pm0.3}$ &
        $45.2$ \\
        & AdaHessian$^\dag$~\cite{AdaHessian} &
        $42.5_{\pm4.8}$ &
        $39.5_{\pm1.0}$ &
        $58.4_{\pm2.6}$ &
        $37.3_{\pm0.8}$ &
        $44.4$ \\
        \hline
        {\multirow{6}*{Sharpness-based}}
        & SAM$^\dag$~\cite{SAM}	&
        $42.9_{\pm3.5}$ &
        $43.0_{\pm2.2}$ &
        $60.5_{\pm1.6}$ &
        $36.4_{\pm1.2}$ &
        $45.7$ \\
        & GAM$^\dag$~\cite{GAM}	&
        $42.2_{\pm2.6}$ &
        $42.9_{\pm1.7}$ &
        $60.2_{\pm1.8}$ &
        $35.5_{\pm0.7}$ &
        $45.2$ \\
        & SAGM*~\cite{SAGM}	&
        $52.2_{\pm3.9}$ &
        $42.3_{\pm0.2}$ &
        $59.9_{\pm0.3}$ &
        $39.9_{\pm2.2}$ &
        $48.6$ \\ 
        & FAD$^\dag$~\cite{FAD}	&
        $44.3_{\pm2.2}$ &
        $43.5_{\pm1.7}$ &
        $60.9_{\pm2.0}$ &
        $34.1_{\pm0.5}$ &
        $45.7$ \\
        & CRSAM*~\cite{CRSAM} &
        $50.8_{\pm0.5}$ &
        $35.3_{\pm1.1}$ &
        $56.8_{\pm0.1}$ &
        $38.4_{\pm1.4}$ &
        $45.3$ \\ 
        & FSAM*~\cite{FSAM}	&
        $47.7_{\pm3.1}$ &
        $39.3_{\pm3.6}$ &
        $57.8_{\pm1.7}$ &
        $39.5_{\pm2.1}$ &
        $46.1$ \\ 
        \hline
        \multicolumn{2}{c|}{MeCAM (Ours)} & 
        $55.6_{\pm1.4}$ &
        $40.8_{\pm3.6}$ &
        $58.5_{\pm0.5}$ &
        $40.5_{\pm1.6}$ &
        $48.9$ \\ 
        \Xhline{1pt}
    \end{tabular}
    }
    \label{tab:TerraIncognita2}
\end{table*}


\begin{table*}[!h]
    \caption{Full results ($mean_{\pm std}$) of our MeCAM and existing DG methods calculated across three trials on DomainNet. The results denoted by $\dag$ are taken from~\cite{SAGM} and~\cite{MADG}, while the results marked by $\ddag$ are inherited directly from the original source. 
    }
    \centering
    \resizebox{\textwidth}{!}{
    \begin{tabular}{c|cccccc|c}
        \Xhline{1pt}
        Algorithm &
        Clip &
        Info &
        Paint &
        Quick &
        Real &
        Sketch &
        Average \\
        \hline
        MMD$^\dag$~\cite{MMD} & 
        $32.1_{\pm13.3}$ &
        $11.0_{\pm4.6}$ &
        $26.8_{\pm11.3}$ &
        $8.7_{\pm2.1}$ &
        $32.7_{\pm13.8}$ &
        $28.9_{\pm11.9}$ &
        $23.4$ \\
        Mixstyle$^\dag$~\cite{MixStyle} & 
        $51.9_{\pm0.4}$ &
        $13.3_{\pm0.2}$ &
        $37.0_{\pm0.5}$ &
        $12.3_{\pm0.1}$ &
        $46.1_{\pm0.3}$ &
        $43.4_{\pm0.4}$ &
        $34.0$ \\
        GroupDRO$^\dag$~\cite{GroupDRO} & 
        $47.2_{\pm0.5}$ &
        $17.5_{\pm0.4}$ &
        $33.8_{\pm0.5}$ &
        $9.3_{\pm0.3}$ &
        $51.6_{\pm0.4}$ &
        $40.1_{\pm0.6}$ &
        $33.3$ \\
        IRM$^\dag$~\cite{IRM} & 
        $48.5_{\pm2.8}$ &
        $15.0_{\pm1.5}$ &
        $38.3_{\pm4.3}$ &
        $10.9_{\pm0.5}$ &
        $48.2_{\pm5.2}$ &
        $42.3_{\pm3.1}$ &
        $33.9$ \\
        ARM$^\dag$~\cite{ARM} & 
        $49.7_{\pm0.3}$ &
        $16.3_{\pm0.5}$ &
        $40.9_{\pm1.1}$ &
        $9.4_{\pm0.1}$ &
        $53.4_{\pm0.4}$ &
        $43.5_{\pm0.4}$ &
        $35.5$ \\
        VREx$^\dag$~\cite{VREx} & 
        $47.3_{\pm3.5}$ &
        $16.0_{\pm1.5}$ &
        $35.8_{\pm4.6}$ &
        $10.9_{\pm0.3}$ &
        $49.6_{\pm4.9}$ &
        $42.0_{\pm3.0}$ &
        $33.6$ \\
        AND-mask$^\ddag$~\cite{SAND-mask} &
        $52.3_{\pm0.8}$ &
        $17.3_{\pm0.5}$ &
        $43.7_{\pm1.1}$ &
        $12.3_{\pm0.4}$ &
        $55.8_{\pm0.4}$ &
        $46.1_{\pm0.8}$ &
        $37.9$ \\
        CDANN$^\dag$~\cite{CDANN} & 
        $54.6_{\pm0.4}$ &
        $17.3_{\pm0.1}$ &
        $43.7_{\pm0.9}$ &
        $12.1_{\pm0.7}$ &
        $56.2_{\pm0.4}$ &
        $45.9_{\pm0.5}$ &
        $38.3$ \\
        SAND-mask$^\ddag$~\cite{SAND-mask} & 
        $43.8_{\pm1.3}$ &
        $15.2_{\pm0.2}$ &
        $38.2_{\pm0.6}$ &
        $9.0_{\pm0.2}$ &
        $47.1_{\pm1.1}$ &
        $39.9_{\pm0.6}$ &
        $32.2$ \\
        DANN$^\dag$~\cite{DANN} & 
        $53.1_{\pm0.2}$ &
        $18.3_{\pm0.1}$ &
        $44.2_{\pm0.7}$ &
        $11.8_{\pm0.1}$ &
        $55.5_{\pm0.4}$ &
        $46.8_{\pm0.6}$ &
        $38.3$ \\
        MTL$^\dag$~\cite{MTL} & 
        $57.9_{\pm0.5}$ &
        $18.5_{\pm0.4}$ &
        $46.0_{\pm0.1}$ &
        $12.5_{\pm0.1}$ &
        $59.5_{\pm0.3}$ &
        $49.2_{\pm0.1}$ &
        $40.6$ \\
        Mixup$^\dag$~\cite{Mixup} & 
        $55.7_{\pm0.3}$ &
        $18.5_{\pm0.5}$ &
        $44.3_{\pm0.5}$ &
        $12.5_{\pm0.4}$ &
        $55.8_{\pm0.3}$ &
        $48.2_{\pm0.5}$ &
        $39.2$ \\
        MLDG$^\dag$~\cite{MLDG} & 
        $59.1_{\pm0.2}$ &
        $19.1_{\pm0.3}$ &
        $45.8_{\pm0.7}$ &
        $13.4_{\pm0.3}$ &
        $59.6_{\pm0.2}$ &
        $50.2_{\pm0.4}$ &
        $41.2$ \\
        ERM$^\dag$~\cite{ERM} & 
        $62.8_{\pm0.4}$ &
        $20.2_{\pm0.3}$ &
        $50.3_{\pm0.3}$ &
        $13.7_{\pm0.5}$ &
        $63.7_{\pm0.2}$ &
        $52.1_{\pm0.5}$ &
        $43.8$ \\
        SagNet$^\dag$~\cite{SagNet} & 
        $57.7_{\pm0.3}$ &
        $19.0_{\pm0.2}$ &
        $45.3_{\pm0.3}$ &
        $12.7_{\pm0.5}$ &
        $58.1_{\pm0.5}$ &
        $48.8_{\pm0.2}$ &
        $40.3$ \\
        Fishr$^\dag$~\cite{Fishr} &
        $58.3_{\pm0.5}$ &
        $20.2_{\pm0.2}$ &
        $47.9_{\pm0.2}$ &
        $13.6_{\pm0.3}$ &
        $60.5_{\pm0.3}$ &
        $50.5_{\pm0.3}$ &
        $41.8$ \\
        MADG$^\dag$~\cite{MADG} &
        $62.5_{\pm0.4}$ &
        $22.0_{\pm0.3}$ &
        $34.1_{\pm0.3}$ &
        $15.1_{\pm0.2}$ &
        $57.4_{\pm1.1}$ &
        $48.0_{\pm0.3}$ &
        $39.9$ \\
        \hline
        MeCAM (Ours) & 
        $64.2_{\pm0.2}$ &
        $21.5_{\pm0.4}$ &
        $51.6_{\pm0.2}$ &
        $14.2_{\pm0.6}$ &
        $63.9_{\pm0.1}$ &
        $54.6_{\pm0.3}$ &
        $45.0$ \\
        \Xhline{1pt}
    \end{tabular}
    }
    \label{tab:DomainNet}
\end{table*}

\begin{table*}[!h]
    \caption{Full results ($mean_{\pm std}$) of our MeCAM, optimizers, and existing sharpness-based DG methods calculated across three trials on DomainNet. Results marked with $\dag$ and * are inherited from~\cite{FAD} and obtained by reproduction, respectively.
    }
    \centering
    \resizebox{\textwidth}{!}{
    \begin{tabular}{c|c|cccccc|c}
        \Xhline{1pt}
        \multicolumn{2}{c|}{Algorithm} & 
        Clip &
        Info &
        Paint &
        Quick &
        Real &
        Sketch &   
        Average \\
        \hline
        {\multirow{6}*{Optimizer}}
        & Adam$^\dag$~\cite{Adam} &
        $63.0_{\pm0.3}$ &
        $20.2_{\pm0.4}$ &
        $49.1_{\pm0.1}$ &
        $13.0_{\pm0.3}$ &
        $62.0_{\pm0.4}$ &
        $50.7_{\pm0.1}$ &
        $43.0$ \\
        & AdamW$^\dag$~\cite{AdamW} &
        $63.0_{\pm0.6}$ &
        $20.6_{\pm0.2}$ &
        $49.6_{\pm0.0}$ &
        $13.0_{\pm0.2}$ &
        $63.6_{\pm0.2}$ &
        $50.4_{\pm0.1}$ &
        $43.4$ \\
        & SGD$^\dag$~\cite{SGD}	&
        $61.3_{\pm0.2}$ &
        $20.4_{\pm0.2}$ &
        $49.4_{\pm0.2}$ &
        $12.6_{\pm0.1}$ &
        $65.7_{\pm0.0}$ &
        $49.6_{\pm0.2}$ &
        $43.2$ \\
        & YOGI$^\dag$~\cite{YOGI} &
        $63.3_{\pm0.1}$ &
        $20.6_{\pm0.1}$ &
        $50.1_{\pm0.3}$ &
        $13.2_{\pm0.3}$ &
        $62.8_{\pm0.1}$ &
        $51.0_{\pm0.2}$ &
        $43.5$ \\
        & AdaBelief$^\dag$~\cite{AdaBelief}	&
        $63.5_{\pm0.2}$ &
        $20.5_{\pm0.1}$ &
        $50.0_{\pm0.3}$ &
        $13.2_{\pm0.3}$ &
        $63.1_{\pm0.1}$ &
        $50.7_{\pm0.1}$ &
        $43.5$ \\
        & AdaHessian$^\dag$~\cite{AdaHessian} &
        $63.3_{\pm0.2}$ &
        $21.4_{\pm0.1}$ &
        $50.8_{\pm0.3}$ &
        $13.6_{\pm0.1}$ &
        $65.7_{\pm0.1}$ &
        $51.4_{\pm0.2}$ &
        $44.4$ \\
        \hline
        {\multirow{6}*{Sharpness-based}}
        & SAM$^\dag$~\cite{SAM}	&
        $63.3_{\pm0.1}$ &
        $20.3_{\pm0.3}$ &
        $50.0_{\pm0.3}$ &
        $13.6_{\pm0.2}$ &
        $63.6_{\pm0.3}$ &
        $49.6_{\pm0.4}$ &
        $43.4$ \\
        & GAM$^\dag$~\cite{GAM}	&
        $63.0_{\pm0.5}$ &
        $20.2_{\pm0.2}$ &
        $50.3_{\pm0.1}$ &
        $13.2_{\pm0.3}$ &
        $64.5_{\pm0.2}$ &
        $51.6_{\pm0.5}$ &
        $43.8$ \\
        & SAGM*~\cite{SAGM}	&
        $63.6_{\pm0.5}$ &
        $21.6_{\pm0.2}$ &
        $51.4_{\pm0.4}$ &
        $14.2_{\pm0.3}$ &
        $64.2_{\pm0.2}$ &
        $53.1_{\pm0.2}$ &
        $44.7$ \\ 
        & FAD$^\dag$~\cite{FAD}	&
        $64.1_{\pm0.3}$ &
        $21.9_{\pm0.2}$ &
        $50.6_{\pm0.3}$ &
        $14.2_{\pm0.4}$ &
        $63.6_{\pm0.1}$ &
        $52.2_{\pm0.2}$ &
        $44.4$ \\
        & CRSAM*~\cite{CRSAM} &
        $62.9_{\pm0.3}$ &
        $20.7_{\pm0.2}$ &
        $50.8_{\pm0.2}$ &
        $15.2_{\pm0.4}$ &
        $62.5_{\pm0.4}$ &
        $53.5_{\pm0.2}$ &
        $44.3$ \\ 
        & FSAM*~\cite{FSAM}	&
        $63.3_{\pm0.5}$ &
        $21.5_{\pm0.1}$ &
        $51.6_{\pm0.3}$ &
        $15.4_{\pm0.5}$ &
        $63.5_{\pm0.2}$ &
        $54.3_{\pm0.2}$ &
        $44.9$ \\ 
        \hline
        \multicolumn{2}{c|}{MeCAM (Ours)} & 
        $64.2_{\pm0.2}$ &
        $21.5_{\pm0.4}$ &
        $51.6_{\pm0.2}$ &
        $14.2_{\pm0.6}$ &
        $63.9_{\pm0.1}$ &
        $54.6_{\pm0.3}$ &
        $45.0$ \\
        \Xhline{1pt}
    \end{tabular}
    }
    \label{tab:DomainNet2}
\end{table*}

\section{Additional Experiments}
\label{Experiment}

\begin{figure*}[!htb]
  \centering
  \includegraphics[width=\textwidth]{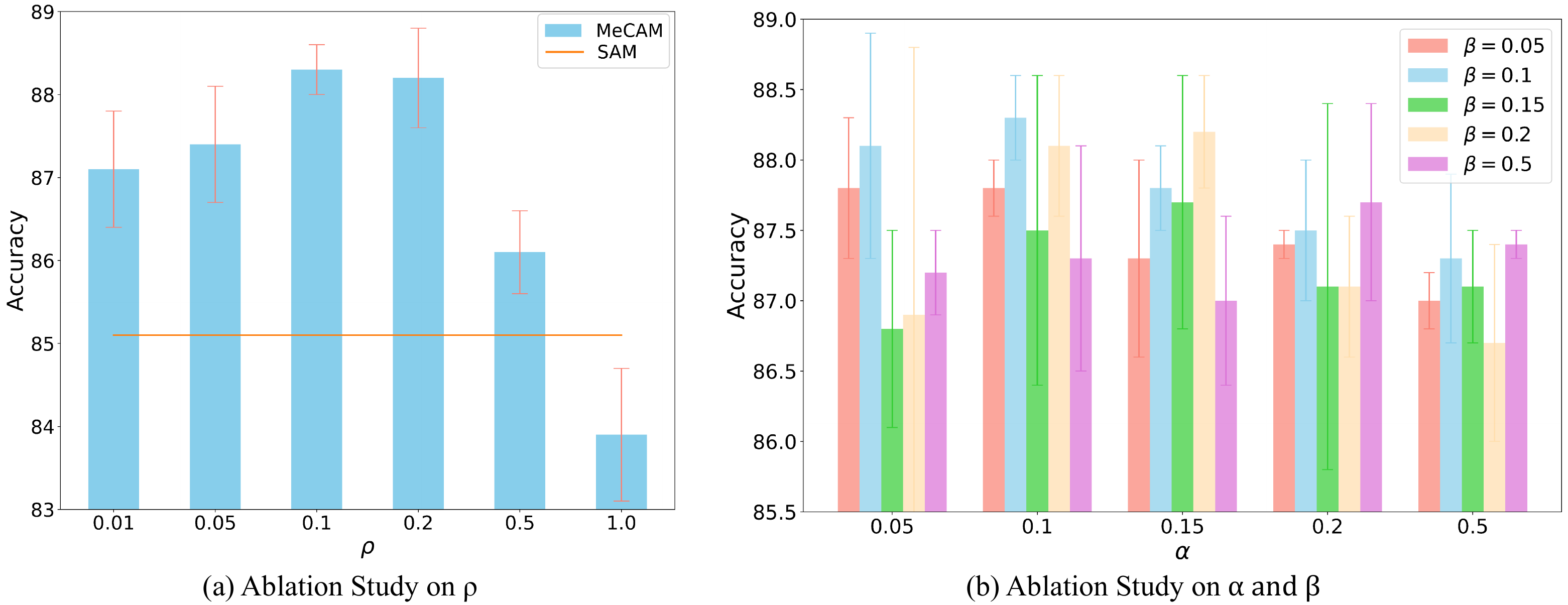}
   \caption{Accuracy of our MeCAM with various $\rho$, $\alpha$, and $\beta$ on the PACS dataset. (a) We evaluated our MeCAM using different $\rho$. (b) We conducted the grid search for $\alpha$ and $\beta$.} 
\label{fig:hyper}
\end{figure*}


\subsection{Discussion on Hyperparameters}
Besides the vanilla training loss, the overall objective function of our MeCAM consists of the surrogate gap of SAM and the surrogate gap of meta-learning. There are three critical hyperpamaters in MeCAM, where $\rho$ controls the extent of perturbations and $\alpha$ and $\beta$ regulate the contributions of the surrogate gap of SAM and meta-learning, respectively.
To explore the impact of these hyperparameters on MeCAM, we conducted additional ablation study on them, and the results are shown in Figure~\ref{fig:hyper}.
Figure~\ref{fig:hyper} (a) shows that (1) the performance remains stable even with variations in $\rho$ around $0.1$; and (2) our MeCAM consistently outperforms SAM, with the exception of the case when $\rho=1.0$, which leads to large perturbations.
The results displayed in Figure~\ref{fig:hyper} (b) indicate that (1) our MeCAM is robust to various $\alpha$ and $\beta$, consistently achieving superior accuracy compared to SAM, even under the worst-case scenario; (2) setting $\beta$ to 0.1 is a proper choice, as it yields the best or second-best performance across different values of $\alpha$; and (3) setting both $\alpha$ and $\beta$ to 0.1 leads to the best overall performance on PACS dataset with a competitive standard deviation.

\end{document}